\documentclass[runningheads]{llncs}

\usepackage{accv}

\usepackage{accvabbrv}

\usepackage{graphicx}
\usepackage{booktabs}

\usepackage[accsupp]{axessibility}  % Improves PDF readability for those with disabilities.

\usepackage[pagebackref,breaklinks,colorlinks,citecolor=accvblue]{hyperref}
\usepackage{orcidlink}

\usepackage[dvipsnames]{xcolor}

\newcommand{\knote}[1]{{\bf \color{magenta}[K:]#1}}
\renewcommand{\knote}[1]{#1}

\usepackage{comment}
\usepackage{url}
\usepackage{bm}
\usepackage{here}
\usepackage{multirow}
\usepackage{amsmath}

\begin{document}

% ---------------------------------------------------------------
\title{Neural Active Structure-from-Motion\\ in Dark and Textureless Environment} 


\author{Kazuto Ichimaru\inst{1,2}\orcidlink{0000-0002-5983-5972} \and
Diego Thomas\inst{1}\orcidlink{0000-0002-8525-7133} \and
Takafumi Iwaguchi\inst{1}\orcidlink{0000-0001-9811-0993} \and \\
Hiroshi Kawasaki\inst{1}\orcidlink{0000-0001-5825-6066}}

\authorrunning{K.~Ichimaru et al.}

\institute{Kyushu University, Japan\\
\url{https://www.cvg.ait.kyushu-u.ac.jp/} \and
Fujitsu Defense \& National Security Limited, Japan\\
\url{https://www.fujitsu.com/jp/group/fdns/}}

\maketitle

\begin{abstract}
Active 3D measurement, especially structured light (SL) has been widely used in various fields for its robustness against textureless or equivalent surfaces by low light illumination.
In addition, reconstruction of large scenes by moving the SL system has become popular, 
however, there have been few practical techniques to obtain the system's precise pose information only from images, since most conventional techniques are based on image features, which cannot be retrieved under textureless environments.
In this paper, we propose a simultaneous shape reconstruction and pose estimation technique for SL systems from an image set where sparsely projected patterns onto the scene are observed (\ie no scene texture information), which we call \textbf{Active SfM}.
To achieve this, we propose a full optimization framework of the volumetric shape that employs neural signed distance fields (Neural-SDF) for SL
with the goal of not only reconstructing the scene shape but also estimating the poses for each motion of the system.
Experimental results show that the proposed method is able to achieve accurate shape reconstruction as well as pose estimation from images where only projected patterns are observed. 
\end{abstract}

\section{Introduction}
\label{sec:intro}

For decades, active 3D measurement has been widely used in the field of autonomous vehicle control, human body analysis, industrial inspection, and so on.
Among them, active stereo techniques, typically structured light (SL), have been widely used because of their simple configuration and high accuracy.
%enables accurate and robust 3D shape acquisition compared to passive stereo techniques. %based methods.
%However, when deployed on real-world applications such as autonomous vehicles, the system mainly relies on the textures of the scenes to calibrate its pose.
% These prerequisites make it difficult to explore extreme environments with little ambient illumination, such as the deep sea.
On the other hand, the reconstruction of large scenes by moving a 3D sensor and integrating results has gained considerable attention for AR/VR applications, such as room reconstruction by smartphones or digital map-making for underwater environments. For this purpose, an active measurement system with an external positional sensor, such as an inertial measurement unit (IMU) or a camera, is commonly used\knote{~\cite{zhou2023high,alzuhiri2023imu}}.
Specifically, structure-from-motion (SfM) is a key technique to achieve precise localization of the system only from an image set. % camera.

\begin{figure}[t!]
    \centering
    \includegraphics[width=0.8\linewidth]{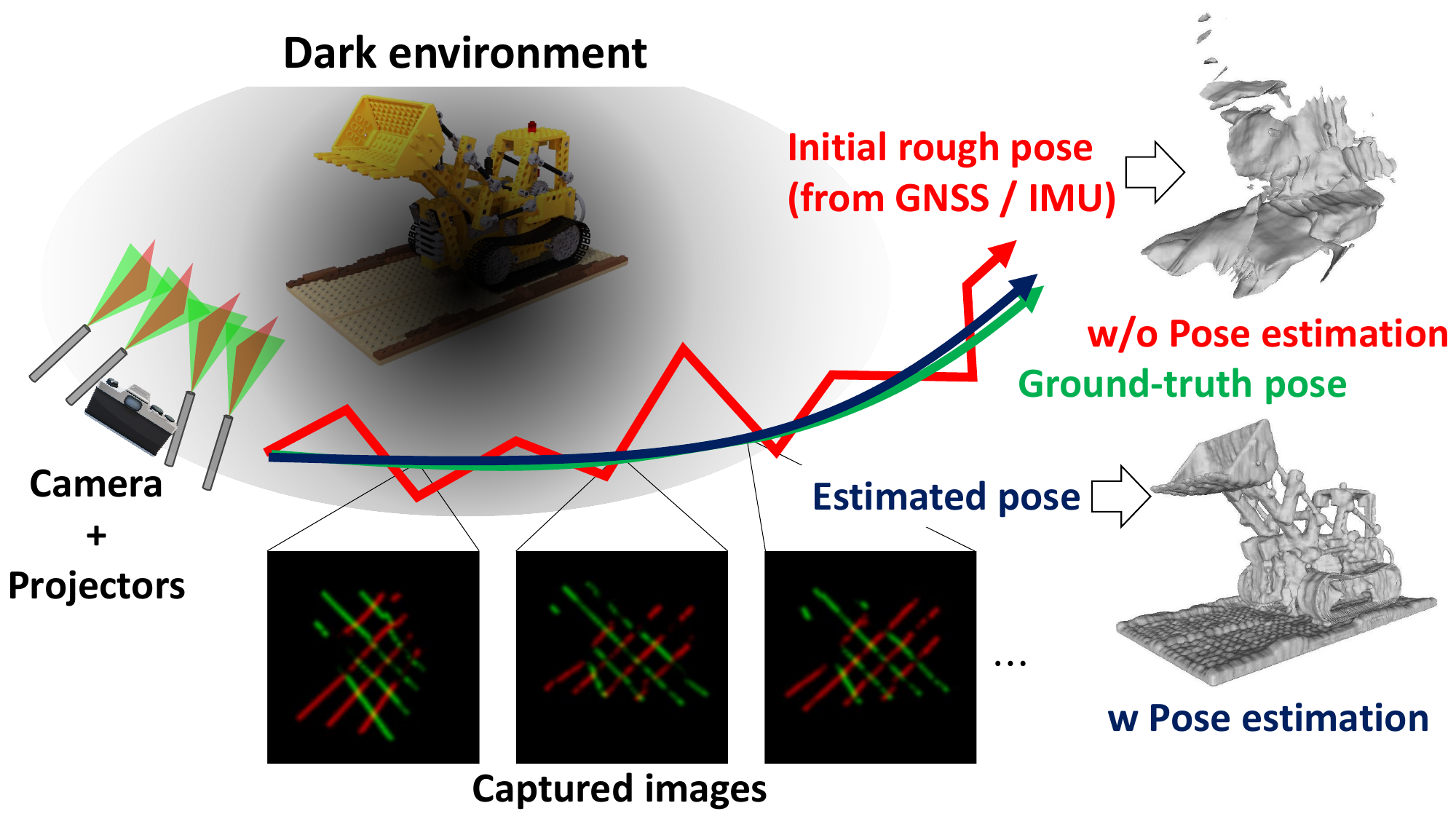}\\
    \caption{Concept of Active Structure-from-Motion (Active SfM). The system consists of a camera and projectors. Images are captured in an extremely dark environment, where texture information is missing. Our goal is to recover the scene shape and the system poses with unreliable initialization from the projected patterns.}
    \label{fig:concept}
%    \vspace{-0.5cm}
\end{figure}

However, it sometimes fails when a room consists of low-texture/uniform walls or during the exploration of extremely dark environments, %with minimal ambient illumination, 
such as the deep sea.
To solve the problem with minimal setup, one may consider using the camera which is used for an active 3D stereo system to estimate the pose of the sensor itself as shown in \autoref{fig:concept}.  

%One fundamental question To overcome the problem, 
%we are interested in 
In this configuration, the fundamental problem is whether it is possible to estimate camera poses \textit{from observed projected patterns by an SL %attached to the camera 
onto the scene}; % as shown in \autoref{fig:concept}.
we name it \textbf{Active Structure-from-Motion (Active SfM)} in the paper.
%The problem, namely \textbf{Active Structure-from-Motion (Active SfM)}, remains untouched due to its difficulty.
Since the area captured by a camera
%where the pattern is projected 
differs frame by frame, if there is no texture or dark environments, it is impossible to retrieve correspondences between frames, making it impossible to apply conventional SfM algorithms.
%to the scene.
%estimate camera poses, which use corresponding points between frames.
%, especially when projecting sparse patterns.
%Therefore, it cannot be solved using conventional methods which use explicit corresponding points based on image features.
Note that even if global navigation satellite system (GNSS) / IMU or other sensors can be used to obtain rough sensor poses, their accuracy is generally insufficient for precise shape integration. %reconstruction.

Recently, Neural Radiance Fields (NeRF)~\cite{NeRF} and its variants have drawn wide attention and brought breakthroughs into many computer vision tasks. %problems. %applications.
By directly optimizing deep neural networks (DNN) to minimize a photometric loss in an end-to-end manner, they achieve remarkable accuracy on novel-view synthesis, 3D-shape reconstruction~\cite{NeuS,neuralangelo}, super resolution~\cite{wang2021nerf-sr}, etc. %and so on.
Some also tried to integrate pattern projection into the NeRF pipeline to introduce neural fields for SL systems~\cite{Chunyu2022DRSL,NFSL}, however, they assume precise pre-calibration and dense reconstruction, which cannot be assumed for the sensor in motion.

In this paper, we propose a novel method to solve the Active SfM using a sparse SL system based on a NeRF variant under dark or even no illumination.
Specifically, we propose a neural signed distance fields (Neural-SDF) that simultaneously estimates the 3D shape of the scene and the poses of the moving camera from the observed patterns projected by the SL system with unreliable initial poses.
The technique is based on a novel volumetric rendering pipeline and hybrid encoding specialized for SL.
Thanks to those proposed methods, it works in a scene where there is little texture or, in the most extreme case, no ambient illumination at all.
Experiments were conducted to prove that the proposed method can solve the Active SfM from projected patterns for both synthetic and real data.
Our contributions are as follows:
\begin{itemize}
    \item We propose a novel Neural-SDF pipeline for Active SfM which enables both shape reconstruction and pose estimation of the SL system in motion from the projected pattern of SL and unreliable initial poses. 
    \item In pursuit of fidelity and robustness, a volumetric rendering pipeline as well as a hybrid encoding technique for SL are proposed.
%    \item The proposed method achieves high accuracy and fidelity thanks to implementation-level optimization.
    \item Comprehensive experiments with both synthetic and real data were conducted to show the feasibility and effectiveness of the proposed method.
\end{itemize}

%\vspace{-0.5cm}
\section{Related work}
\label{sec:related_work}

\subsection{Neural Radiance Fields}
Neural Radiance Fields (NeRF) is a methodology to represent a scene as a volumetric function that outputs the density of a 3D point and color from a multiple views~\cite{NeRF}.
%specific viewing direction~\cite{NeRF}.
NeRF utilizes DNN's ability of interpolation and extrapolation to accurately generate novel views from limited images, or estimate scene shape.
%Some use NeRF as a BRDF tool for albedo acquisition or re-lighting~\cite{NeRD}.
While NeRF performs well on novel view synthesis, its performance on 3D-shape reconstruction is limited, since volumetric density function is not suitable to represent solid surfaces.
VolSDF and NeuS replaced volumetric density function with signed distance function (SDF), which outputs signed distance to the closest surface~\cite{VolSDF,NeuS,NeuS2}.
Neural SDF drastically improved 3D-shape reconstruction accuracy by introducing inductive bias.
%\knote{
Although some density function based methods achieved highly accurate reconstruction~\cite{NeRFMeshing}, SDF is suitable for various geometric down-stream tasks, such as normal reguralization by Eikonal loss~\cite{Eikonal}, shape integration by TSDF~\cite{KinectFusion}, shape editing~\cite{tzathas2023}, and so on. %}
%\vspace{-0.25cm}

\subsection{Structure-from-Motion}
Structure-from-Motion (SfM) is a task to simultaneously estimate scene shape and camera motion.
While conventional SfM methods based on image features have been successful~\cite{schoenberger2016sfm}, learning-based SfM and Differentiable Rendering (DR)-based SfM have been proposed in pursuit of higher accuracy and robustness~\cite{SFMLearner,DeepV2D}.
NeRF is also known that it can efficiently solve SfM-like problems.
Yen-Chen \etal used NeRF as a pose estimator from known shape and rendering.
Later, some found simultaneous shape reconstruction and pose estimation is possible~\cite{NeRF--,BARF,SCNeRF,CamP,Zhang_2023_ICCV}.
Wang \etal also tried to estimate intrinsic parameters such as focal length~\cite{NeRF--}, while Lin \etal proposed dedicated positional encoding to enhance the robustness of pose estimation~\cite{BARF}.
However, none of them achieved shape reconstruction and pose estimation without texture information to the best of our knowledge.

\knote{
One major challenge of SfM is robustness against dark and textureless environments.
Conventional SfM uses image feature based localization, which is very challenging in such environments as shown in \autoref{fig:icp_failure} (Right).
There are several NeRF methods that achieve NVS in dark environments, but they often require special image formats such as raw images or metadata~\cite{mildenhall2022rawnerf,huang2022hdr}.
Recently, LLNeRF achieved robust NVS under dark illumination environment without special image format, which is evaluated in the experiments~\cite{llnerf}.
Furthermore, there is no method that achieves pose estimation in such an environment, to the best of our knowledge.
}

\subsection{Active Stereo for SfM}
%\vspace{-0.25cm}
Active stereo is a methodology to estimate scene shape using active light sources.
In a broad sense, active stereo may include projected pattern stereo~\cite{ProjectedTextureStereo} or photometric stereo~\cite{ActivePhotometricStereo}, but we focus on structured light (SL) in this paper because of its simplicity and robustness. %context.
SL uses projected patterns as clues to find correspondences for triangulation.
To eliminate the ambiguity of correspondences, multiple patterns have been commonly used for SL,
%methods use dedicated patterns, 
such as Graycode~\cite{Graycode}, Phase-shifting~\cite{PhaseShift}.
Recently, investigations into reducing the number of patterns through optimization have been conducted~\cite{Parsa2018cvpr,chen_2020_autotuning}.
%Recently, reducing the number of pattern based on optimization are also investigated~\cite{Parsa2018cvpr,chen_2020_autotuning}.
To apply SL for dynamic scenes, such as SfM, it is preferable to use single pattern, known as oneshot scan, for example, grid-pattern~\cite{Furukawa_2022_WACV}, random dot pattern~\cite{Gu2020sensors}, colorful line pattern~\cite{Fernandez2012icip} and cross laser pattern~\cite{Nagamatsu2021IROS,Nagamatsu2022ICPR}.
%facilitate searching correspondences and 
%reduce number of required images.
%Some SL methods optimize patterns by themselves to minimize ambiguity of correspondences and maximize useful information~\cite{Parsa2018cvpr,chen_2020_autotuning}.
%

\begin{figure}[t!]
    \centering
    \begin{minipage}[b]{0.48\linewidth}
        \centering
        \includegraphics[width=0.8\linewidth]{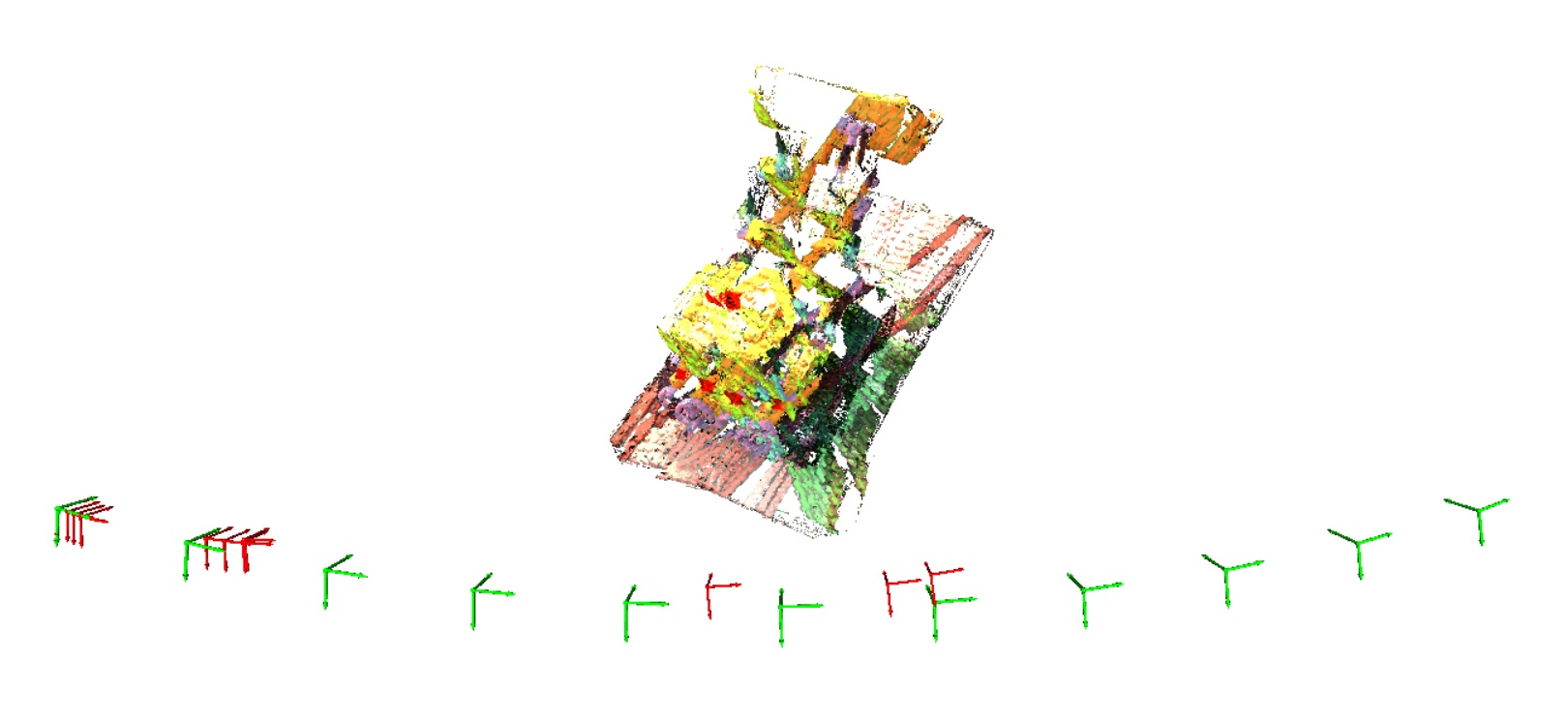}\\
    \end{minipage}
    \begin{minipage}[b]{0.48\linewidth}
        \centering
        \includegraphics[width=\columnwidth]{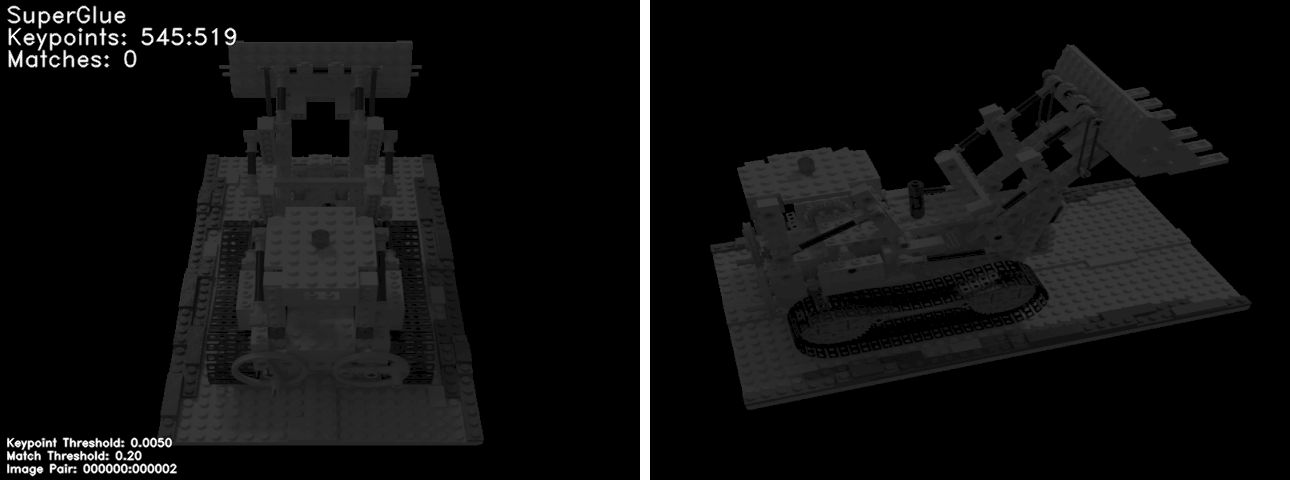}
    \end{minipage}
    \caption{\textbf{Left:} A failure case of ICP-based pose estimation with NeRF-Synthetic (Lego) scene with cross laser projectors. \textbf{Green arrows}: Ground-truth camera poses. \textbf{Red arrows}: Estimated camera poses via ICP with sparsely reconstructed point clouds. Note that the Ground-truth poses are used for initialization.
    \knote{\textbf{Right:} A failure case of SuperGlue~\cite{sarlin20superglue} feature matching with NeRF-Synthetic (Lego) scene with little illumination (No matches are detected). Note that the contrast is enhanced for visualization.}}
    \label{fig:icp_failure}
%    \vspace{-0.5cm}
\end{figure}

To reconstruct the shapes of large-scale scenes or the entire shape of objects, it is possible to scan the scene with a one-shot scan followed by shape registration using the iterative closest point (ICP) algorithm~\cite{KinectFusion}.
%To reconstruct the shapes of large scale scenes or entire shape of objects, scanning the scene by oneshot scan followed by shape registration by iterative closest point (ICP) is possible~\cite{KinectFusion}.
%However, in one-shot scan, since positional information are spatially encoded into the pattern, there is trade of between robustness/accuracy and density on shape reconstruction. Therefore, the more accurate we need, the reconstruction results becomes more sparse, making ICP algorithm unstable or even sometimes fails.
%ChatGPT
However, in a one-shot scan, as positional information is spatially encoded into the pattern, a trade-off arises between robustness/accuracy and density in shape reconstruction. Therefore, the pursuit of greater accuracy leads to sparser reconstruction, potentially resulting in ICP algorithm unstable or even causing occasional failures as demonstrated in \autoref{fig:concept} and \autoref{fig:icp_failure} (Left).
Some work used multiple cross-laser projectors instead of an ordinary video projector to achieve simultaneous dense shape reconstruction and self-calibration of the system~\cite{Kawasaki2009:Article_Laser1276714555,Nagamatsu2022ICPR}.
In \cite{Nagamatsu2022ICPR}, it is said that the problem has essentially 4DOF indeterminacy and extra constraints are required to solve the problem, proposing to use 4 cross-laser projectors, and we follow their system configuration.
However, \cite{Nagamatsu2022ICPR} uses pre-trained laser detection model, which requires labor-some annotation.

%There are also commercial depth cameras using SL methods~\cite{Kinect,RealSense}.

%Recently, Differentiable Rendering (DR) has drawn wide attention for its effectiveness in end-to-end optimization.
%Some use DR to optimize pattern~\cite{baek2021polka}, while others use DR as inverse problem solvers to directly estimate scene as depth image~\cite{Benjamin2021DDS}, or triangle mesh~\cite{Janus2021SLDR}.
%DR is also closely related to NeRF, which is mentioned in the next section.

Recently, NeRF pipelines for SL systems have been proposed to incorporate into NeRF the robustness of SL systems against insufficient scene texture and illuminance.
Li \etal combined Graycode SL~\cite{Graycode} with Neural SDF to cope with an inter-reflection problem, which severely degrades reconstruction quality~\cite{Chunyu2022DRSL}.
Shandilya \etal also integrated a Random dot pattern projector into the NeRF pipeline and proposed a model to separate ambient light, direct illumination, and indirect illumination from the light source~\cite{NFSL}.
Inspired by the current progress of NeRF / Neural SDF techniques which enabled simultaneous shape reconstruction and camera pose estimation, we propose a Neural SDF pipeline for SfM with SL,
including extremely sparse patterns such as several number of line-lasers.

\ifx
Although SL has made great progress in pursuit of high accuracy and robustness for in-the-wild scenes, there is no practical method that has succeeded in estimating camera poses based on the projected patterns (Active SfM), as far as we know.
For a specific device, Iterative Closest Point (ICP)-based pose estimation is possible~\cite{KinectFusion}, but ICP may lead to suboptimal results when dense pattern is not available as demonstrated in \autoref{fig:icp_failure}.
Inspired by the current progress of NeRF / Neural SDF techniques which enabled simultaneous shape reconstruction and camera pose estimation, we propose a Neural SDF pipeline for Active SfM with arbitrary patters, including extremely sparse ones such as lasers.
\fi

%\vspace{-0.25cm}

\section{Method Overview}
\label{sec:method_overview}

%In this section, we describe the overview of the proposed method.

\begin{figure}[t!]
    \centering
    \includegraphics[width=0.95\linewidth]{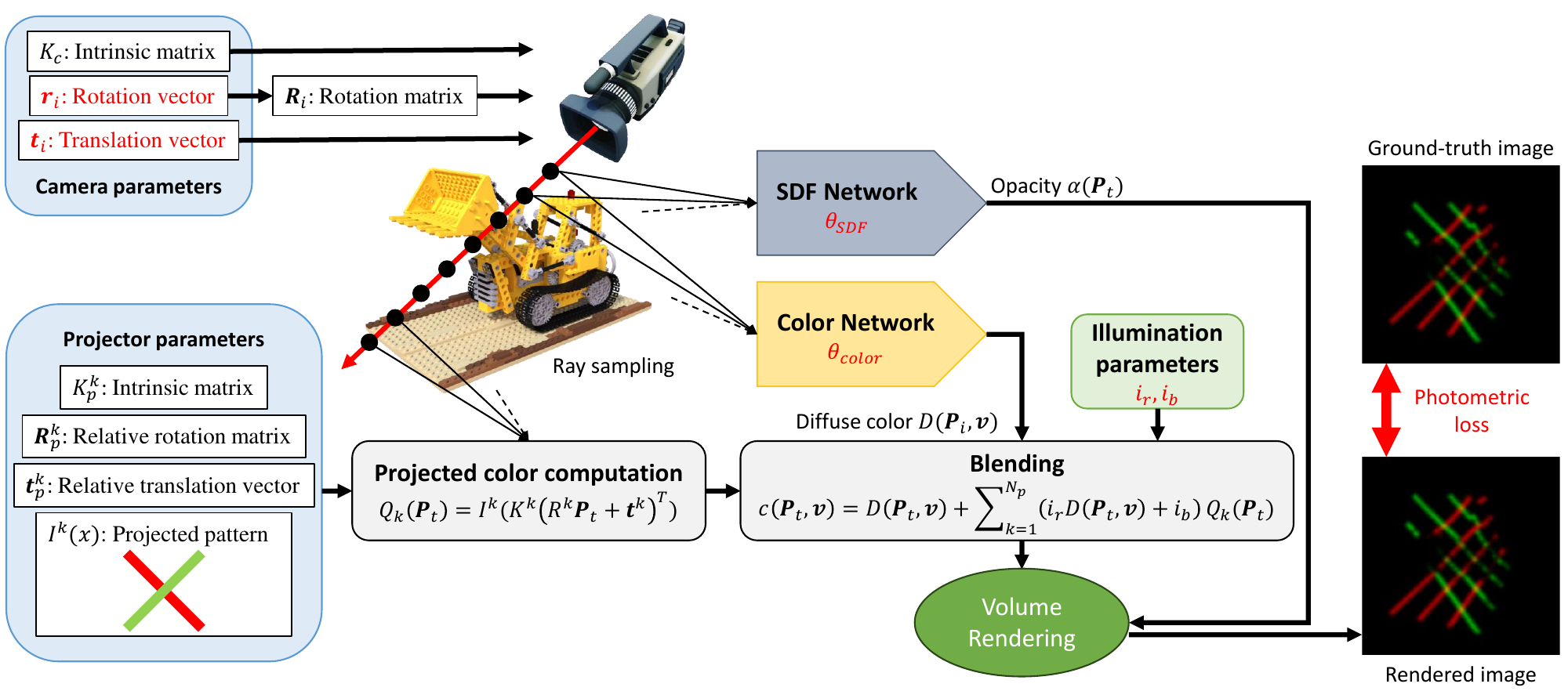}\\
    \caption{Pipeline schematics of the proposed method. Parameters marked in red are optimized during training.}
    \label{fig:pipeline}
%    \vspace{-0.5cm}
\end{figure}

%\vspace{-0.25cm}

\subsection{System configuration and environmental assumption}

The proposed method assumes system configuration with a camera and an arbitrary number of projectors whose intrinsic parameters are accurately calibrated, as shown in \autoref{fig:concept}.
Relative transformations between the camera and the projectors are fixed and calibrated during the scan. %capturing.
We assume rough system poses are available for initialization from external devices such as GNSS or IMU, but their accuracy is insufficient for shape reconstruction and integration as shown in the experiments.
%The projectors project arbitrary patterns, such as Graycode, Grid-pattern, Cross-laser pattern, and so on.
%The patterns do neither have to be static during capturing, nor consistent across the projectors in theory, but we used static and consistent patterns (mainly Cross-laser pattern) in the experiments.
The projectors project arbitrary patterns including very sparse patterns such as cross-line-lasers, which is used in the experiment.
As for the environment, we assume the scenes are static during scan, and Lambertian reflectance is dominant on material surfaces.
%\vspace{-0.25cm}

\subsection{Pipeline of the proposed method}

The proposed method mainly follows NeuS~\cite{NeuS} pipeline, which consists of the SDF network and the color network, except that our pipeline has \textbf{Projector parameters}, \textbf{Projected patterns} and  \textbf{Illumination parameters}.% which consist \textbf{Volumetric rendering with pattern projection} branch.
\autoref{fig:pipeline} shows the pipeline of the proposed method.
Note that we omit some modules from the figure used in the pipeline, such as hierarchical sampling and variance network, which are also used by NeuS.

The training procedure of the pipeline is as follows.
\knote{
Bold lines are the processes for the proposed method, while others are almost identical to common NeRF pipeline.
}
%Note that procedures 1-3 and 7,8 are almost identical to that of NeuS.
\begin{enumerate}
    \item Randomly sample rays casting from each camera's optical centers using camera parameters.
    \item Sample 3D points on the ray from the near clip to the far clip at regular or weighted intervals.
    \item Pass the 3D points to the SDF / color networks to acquire the density and albedo of the points.
%    \item Re-project the 3D points onto the projector patterns to get 2D projected points using projector parameters.
%    \item Compute projected colors to the 3D points from the projected patterns using the 2D projected points.
    \item \textbf{Render images by volumetric rendering with pattern projection}.
    \item Compute photometric loss between the rendered images and the ground-truth (GT) images.
    \item \textbf{Update the network parameters and the system poses to minimize the photometric loss using Adam}.
\end{enumerate}

By updating the network parameters, SDF is optimized to minimize the discrepancy between the projected pattern and the GT image, \ie implicitly searching image-pattern correspondences and computing scene depth from the views.
At the same time, system poses are optimized to maximize multi-view depth consistency.
%We assume texture information is missing as the most challenging scenario, but the pipeline enables implicit utilization of texture information when it is available to learn more detailed SDF as ordinary NeRF.

%\vspace{-0.3cm}

\section{Neural SDF for Active SfM}
\label{sec:method}

%In this section, we describe the details of the proposed method.

\subsection{Ray sampling}

%As for the process of ray sampling, we follow NeuS implementation with a few modifications.
Given an image, we randomly sample $N$ pixel coordinates $(p_x, p_y)$, and convert to homogeneous coordinates $\bm{p} = (p_x, p_y, 1)$.
\knote{
Next, we backcast the coordinates to compute the ray direction vector $\bm{P}$ and ray origin $\bm{O}$ in world coordinate system using the camera's intrinsic matrix $K_c$, $i$-th system rotation matrix $\bm{R}_i$ and translation vector $\bm{t}_i$ as follows:
\begin{equation}
    \begin{split}
        \bm{P} &= \bm{R}_i \cdot K_c^{-1} \cdot \bm{p}^T, \\
        \bm{O} &= s\bm{t}_i, 
    \end{split}
\end{equation}
where $s$ is the predefined scaling coefficient to fit the entire scene into the unit sphere.
Finally, we sample 3D points along the ray,
\begin{equation}
    \bm{P}_u = \bm{P} \cdot d(u) + \bm{O}
\end{equation}
where $d(u)$ returns a distance from the origin $\bm{O}$.
}
As mentioned below, we implement the ray sampling process in a fully differentiable way to backpropagate photometric loss to the system pose parameters $\bm{R}_i$ and $\bm{t}_i$.

%\vspace{-0.25cm}
\subsection{Volumetric rendering with pattern projection}

In ordinary Neural SDF, a 2D point $p$ is rendered using 3D points $\bm{P}_i$ ($i=1..n$) on the ray cast from $\bm{p}$ as \autoref{eq:volumetric_rendering}
\begin{equation}
    C = \sum_{t=1}^n \left( \prod_{j=1}^{t-1}1-\alpha(\bm{P}_j) \right) \alpha(\bm{P}_t) c(\bm{P}_t, \bm{v}),
    \label{eq:volumetric_rendering}
\end{equation}
where $\alpha(\bm{P}_t)$ is opacity of a 3D point $\bm{P}_t$, $c(\bm{P}_t, \bm{v})$ is color of $\bm{P}_t$ viewing from direction $\bm{v}$, and $C$ is final rendered color of the ray.
Following NeuS, $\alpha(\bm{P}_t)$ is computed as \autoref{eq:alpha},
\begin{equation}
    \alpha(\bm{P}_t) = \max\left(\frac{\Phi_s(f(\bm{P}_t)) - \Phi_s(f(\bm{P}_{t+1}))}{\Phi_s(f(\bm{P}_t))}, 0\right),
    \label{eq:alpha}
\end{equation}
where $f(\bm{P})$ is SDF value of a 3D point $\bm{P}$ and $\Phi_s$ is Sigmoid function in our implementation.

To compute $c(\bm{P}_t,\bm{v})$, the color of a 3D point with pattern projection, we follow a common linear Lambertian model~\cite{NeuralLambertian}.
Specifically, we blend albedo color of $\bm{P}_t$ (denoted as $D(\bm{P}_t,\bm{v})$) and projected color by $k$-th projector (denoted as ${Q_k(\bm{P}_t)}$) as \autoref{eq:color_blend},
\begin{equation}
    c(\bm{P}_t,\bm{v}) = D(\bm{P}_t,\bm{v}) + \sum_{k=1}^{N_p} (i_{r}D(\bm{P}_t,\bm{v}) + i_{b})Q_k(\bm{P}_t),
    \label{eq:color_blend}
\end{equation}
where $i_r$ is reflectance coefficient, and $i_b$ is bias coefficient, integrated as learnable parameters, and $N_p$ is the number of projectors.
$i_r$ controls reflectance level of $\bm{P}_t$, and $i_b$ controls emissive level of $Q(\bm{P}_t)$.
Once $i_r$ and $i_b$ are learned, we can render a high-fidelity image with pattern projection.
% 式を3DVのcopyにならないように変更

%The intuition of this equation is that, if we can render an image with pattern projection correctly, it means we have an accurate depth, texture, and pose, and vice versa.
%Thus, it implicitly works as depth supervision, which helps better 3D shape and pose optimization, especially in texture-less regions or low illumination environments that lack information for depth supervision.

In our implementation, $D(\bm{P}_t,\bm{v})$ is the Color network itself, and $Q_k(\bm{P}_t)$ is computed as \autoref{eq:projector_projection},
\begin{equation}
    Q_k(\bm{P}_t) = I^k(K^k(R^k\bm{P}_t + \bm{t}^k)^T),
    \label{eq:projector_projection}
\end{equation}
where $I^k(x)$ returns the color of the pattern of $k$-th projector at specific point $x$ via bilinear sampling, $K^k$ is the intrinsic matrix and $R^k, \bm{t}^k$ are the relative transformation of $k$-th projector, which convert 3D points in the world coordinate system into the projector screen coordinate system.
%\vspace{-0.25cm}

\subsection{System pose estimation}
\knote{
To estimate the system pose, we define the rendering pipeline as fully differentiable from the system's extrinsic parameters.
The extrinsic parameters consist of rotation and translation (3 degrees of freedom, respectively).
Surprisingly, such a simple modification enables pose estimation under severe conditions, such as extremely low illumination environments.
We define rotations as rotation vectors to remove redundancy and convert them to rotation matrices to compute ray directions, as done in ~\cite{NeRF--}.
}

Note that we assume relative poses between the camera and the projectors are fixed and calibrated in most cases, but they can be also refined by defining them as learnable parameters.
Please refer to the supplementary material for an experiment on projector pose refinement.

%\vspace{-0.25cm}
\subsection{Hybrid encoding for robust and high-fidelity optimization} % and 2-stages training}
It is well-known that positional encoding plays an important role in NeRF-based methods to achieve higher accuracy and fidelity.
While original NeRF and NeuS use Fourier encoding~\cite{NeRF,NeuS}, InstantNGP and NeuS2 drastically improved reconstruction quality by introducing multi-resolution hash encoding~\cite{InstantNGP,NeuS2}, computed as following,
\begin{equation}
    \begin{split}
    e^h_{l}(x) = \sum_{i=1}^{2^d} w_{i,l} \mathcal{H}_l(h(x)), \\
    e^h(x) = \left(h_1(x), h_2(x), \cdots, h_L(x) \right),
    \end{split}
\end{equation}
where $d$ is the dimension of the grids (usually $d=3$), $L$ is the number of resolutions, $\mathcal{H}_l$ is the hash entry of $l$-th resolution, and $h$ is the arbitrary hash function.
Multi-resolution hash encoding enables high-fidelity and extremely fast scene shape reconstruction by independently optimizing parameters for each hash entry.
However, we empirically observed that using multi-resolution hash encoding in our pipeline severely degrades reconstruction and pose estimation quality due to its locality, causing a trade-off between the robustness of pose estimation and fidelity of the reconstructed shape.

To avoid such a phenomenon and still achieve higher accuracy and fidelity, we apply a modification to the positional encoding.
We propose hybrid encoding, \ie, concatenating embedded vectors from Fourier encoding and Multi-resolution hash encoding.
Hybrid encoding $e(x)$ is represented as following,
\begin{equation}
    e(x) = \left(e^f(x), e^h(x) \right),
\end{equation}
where $e^f(x)$ is Fourier encoding, computed as follows,
\begin{equation}
    \begin{split}
    e^f(x) = \left( sin(2^0x), sin(2^1x), \cdots, sin(2^{L-1}x), \right. \\
               \left. cos(2^0x), cos(2^1x), \cdots, cos(2^{L-1}x) \right).
    \end{split}
\end{equation}
Hybrid encoding relaxes the locality of multi-resolution hash encoding and improves reconstruction and pose estimation quality, as shown in the experiments.
%\vspace{-0.5cm}

\subsection{Implementation details}
We implemented the proposed method following NeuS implementation.
We used hierarchical sampling~\cite{NeRF}, Eikonal loss~\cite{Eikonal}, and mask loss~\cite{NeuS} as well for better efficiency and accuracy.
Eikonal loss is a well-known regularizer, which helps in learning a spatially consistent SDF. 
%As for the occlusion of pattern projection, we found that we can just ignore it and let the SDF network learn forward-backward relation thanks to NeuS's occlusion-aware weighting function design.
Overall, the objective function of the pipeline is as \autoref{eq:objective}
\begin{equation}
    \mathcal{L} = \mathcal{L}_{color} + \lambda\mathcal{L}_{reg} + \beta\mathcal{L}_{mask},
    \label{eq:objective}
\end{equation}
where $\mathcal{L}_{color}$ is photometric loss (L1), $\mathcal{L}_{reg}$ is Eikonal term, $\mathcal{L}_{mask}$ is mask loss term, and $\lambda, \beta$ are balancing coefficients.

As for the hyper-parameters of training, we used learning rate $5e-4$, learning rate decay coefficient $0.05$, batch size $512$, Eikonal term coefficient $\lambda = 0.1$, and mask loss term coefficient $\beta = 0.1$.
We trained the networks for 200k steps in the experiments.
%\vspace{-0.3cm}

\section{Experiments}
\label{sec:experiments}

In this section, we describe the experiments with synthetic and real data to confirm the feasibility of the method.

\subsection{Comparative methods}
Through the experiments, we compare the proposed method with the following comparative methods.
Please refer to the supplementary material for the implementation details.
Note that comparison to NeuS, NeuS+Pose estim. and NeuS+SL covers ablation study of the proposed method.
\begin{itemize}
    \item Light-sectioning: A method for active 3D shape reconstruction with sparse set of patterns, typically configured by line-laser projectors.
    \item LLNeRF~\cite{llnerf}: A method for NVS under dark environments.
    \item NeuS~\cite{NeuS} (Ours w/o pose estim. and pattern projection): A method for passive 3D shape reconstruction with SDF.
    \item NeuS+Pose estim. (Ours w/o pattern projection): A NeuS variant with the same pose estimation pipeline to the proposed method.
    \item NeuS+SL (Ours w/o pose estim.): A NeuS variant with structured light, which is identical to the proposed method without pose estimation.
\end{itemize}
%\vspace{-0.5cm}

\subsection{Evaluations with synthetic data}
\label{ssec:synthetic_evaluation}
%\vspace{-0.5cm}

\begin{figure}[tb]
    \centering
    \includegraphics[width=0.35\linewidth]{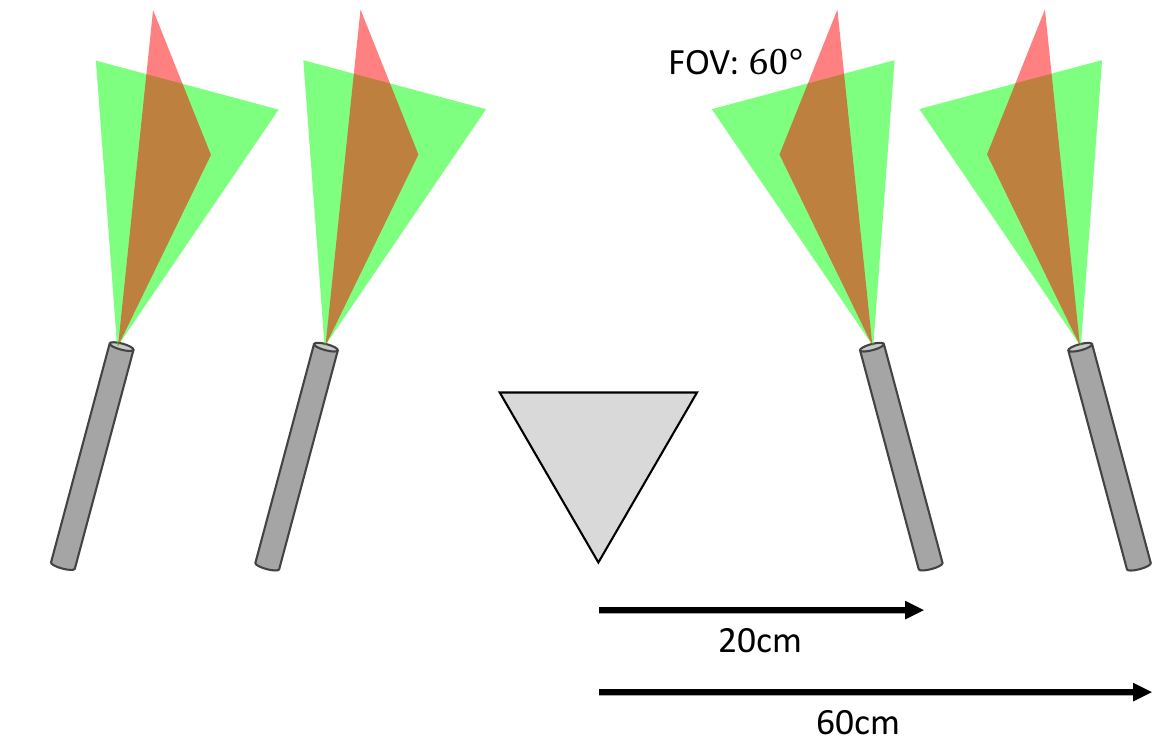}
    \includegraphics[width=0.35\linewidth]{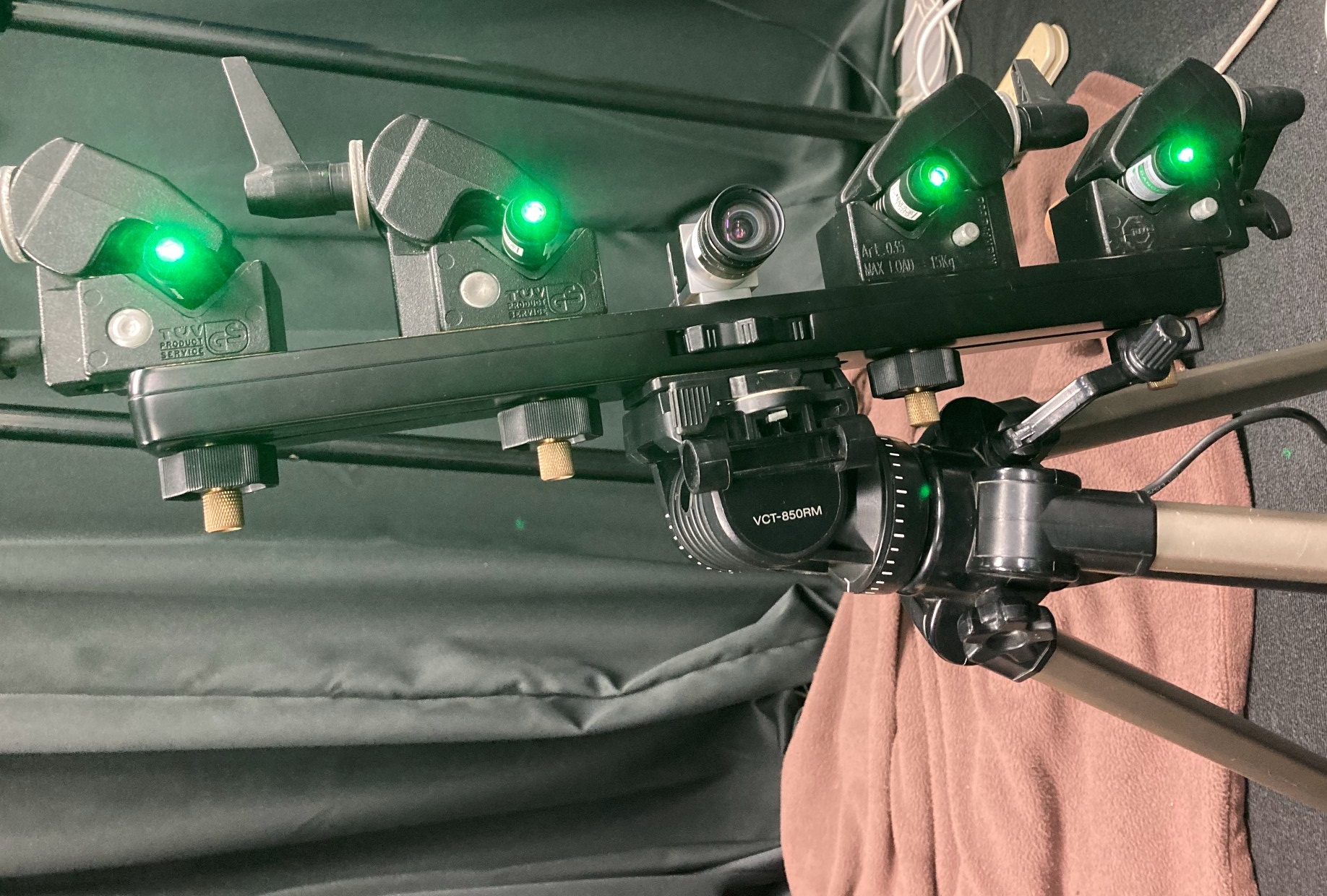}
    \caption{System configuration of SL system in the experiments. Note that the lasers are colored red and green in the left figure to make the configuration understood easily, however, single color is sufficient as can be seen in the real system, all green (right).}% Four cross-line laser projectors are lined up on the left and right of the camera.}
    \label{fig:SL_config}
%    \vspace{-0.5cm}
\end{figure}

\begin{table}[]
    \centering
    \caption{List of scenes of synthetic datasets.}
    \label{tab:datasets}
    \scalebox{0.9}{
    \begin{tabular}{|c|c|c|}
        \hline
%        Dataset & Scenes & \# of images & Baseline scalar \\ \hline \hline

        Dataset & Scenes (\# of images) & Baseline scalar \\ \hline \hline
        NeRF-Synthetic & Lego(40), Chair(40), Hotdog(40), Mic(40) & 1.0 \\ \hline
        BlendedMVS & Stone(56), Dog(31), Bear(123), Sculpture(79) & 0.2 \\ \hline

%        \multirow{4}{*}{NeRF-Synthetic} & Lego & 40 & \multirow{4}{*}{1.0} \\
%        & Chair & 40 & \\
%        & Hotdog & 40 & \\
%        & Mic & 40 & \\ \hline

%        \multirow{4}{*}{BlendedMVS} & Stone & 56 & \multirow{4}{*}{0.2} \\
%        & Dog & 31 & \\
%        & Bear & 123 & \\
%        & Sculpture & 79 & \\
        \hline
    \end{tabular}}
%    \vspace{-0.25cm}
\end{table}

\begin{figure}[t!]
    \centering
    \hspace{-0.3cm}
    \begin{minipage}[b]{0.17\linewidth}
        \centering
        \includegraphics[width=\columnwidth]{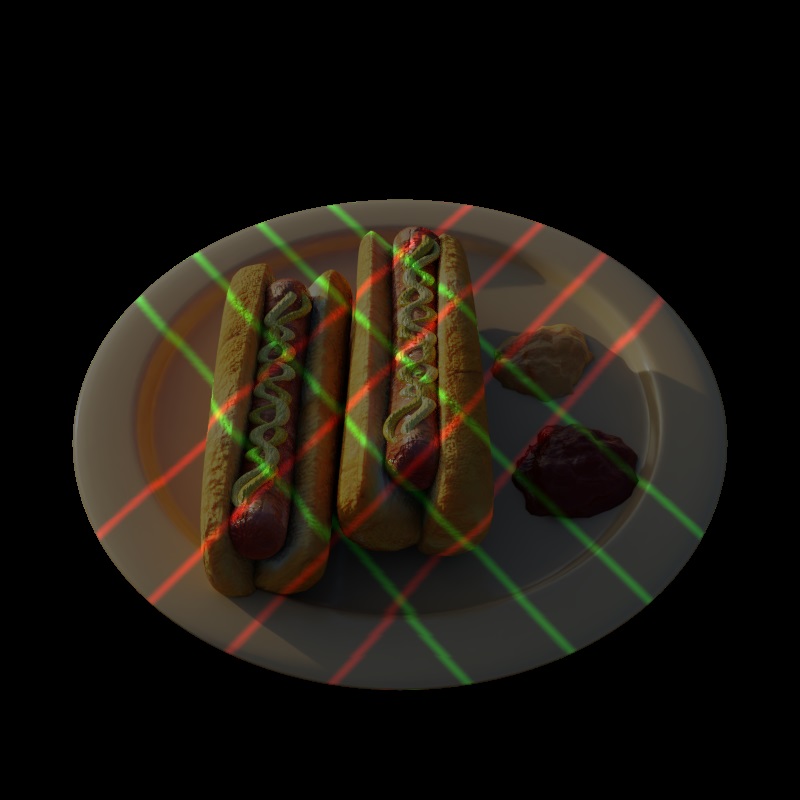}
%        \small Normal illum.
        \footnotesize Normal illum.
    \end{minipage}
    \begin{minipage}[b]{0.17\linewidth}
        \centering
        \includegraphics[width=\columnwidth]{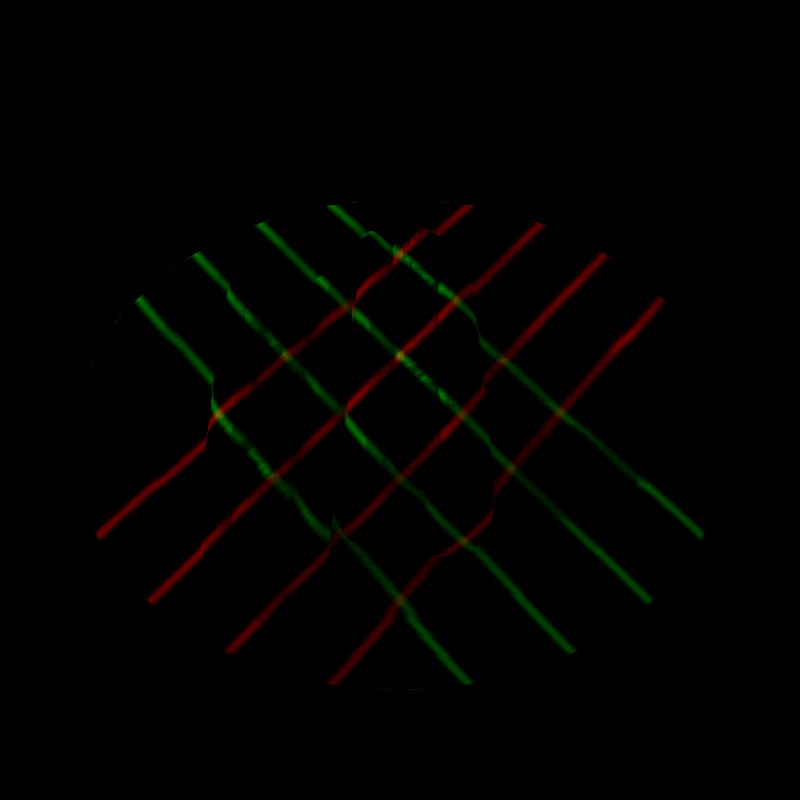}
        \footnotesize No illum.
    \end{minipage}
    \begin{minipage}[b]{0.23\linewidth}
        \centering
        \includegraphics[width=\columnwidth]{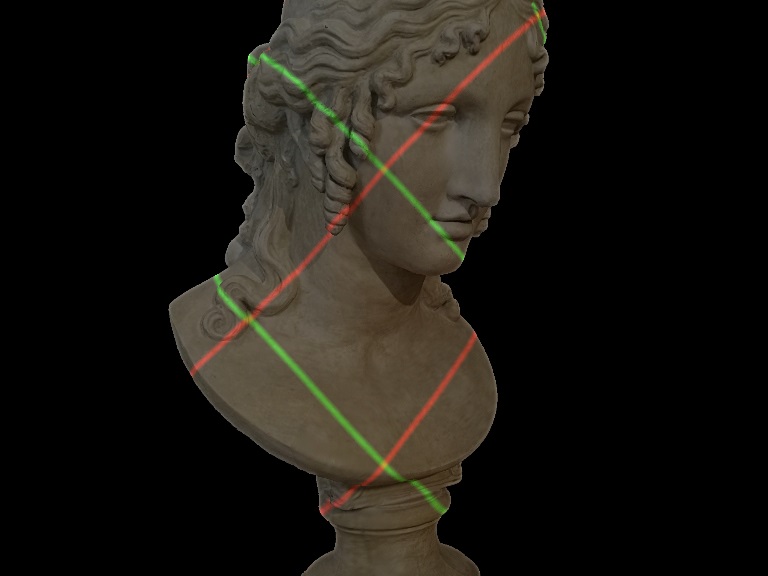}
        \footnotesize Normal illum.        
    \end{minipage}
    \begin{minipage}[b]{0.23\linewidth}
        \centering
        \includegraphics[width=\columnwidth]{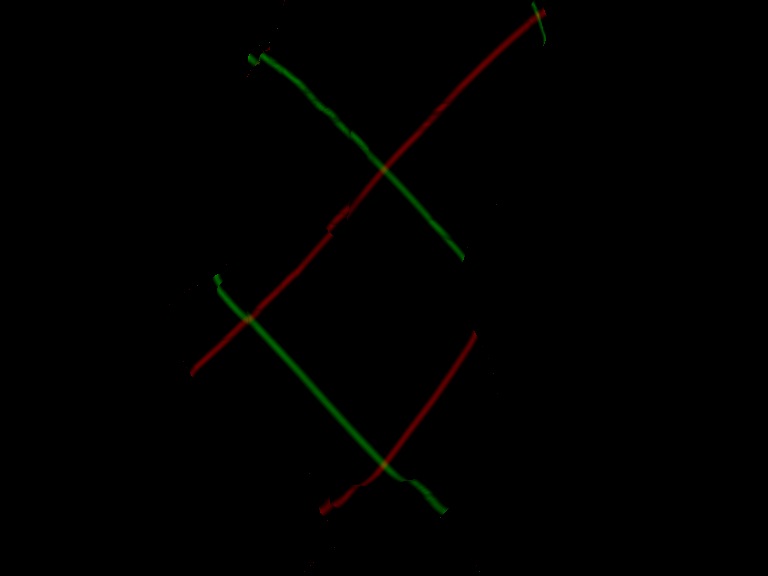}
        \footnotesize No illum.
    \end{minipage}
    \caption{Example images of the synthetic data with pattern projection. \textbf{Left}: NeRF-Synthetic. \textbf{Right}: BlendedMVS.}
    \label{fig:dataset_image}
%    \vspace{-0.25cm}
\end{figure}

We conducted several experiments with synthetic data. % in some respects.
Throughout the experiments, we used two datasets such as NeRF-Synthetic~\cite{NeRF} and BlendedMVS~\cite{yao2020blendedmvs}, as shown in \autoref{tab:datasets}.
\begin{comment}
\begin{itemize}
    \item NeRF-Synthetic\cite{NeRF}: Sequences with a single object captured by a camera moving along an orbital trajectory.
    \item BlendedMVS\cite{yao2020blendedmvs}: Sequences of various scenes including aerial scenery, sculptures, potteries, and so on.
\end{itemize}
\end{comment}
For both datasets, we synthesized images with pattern projection by computing rays from virtual projectors.
We used 4 virtual projectors lined up on the left and right of the camera with static cross-line-laser patterns (red and green) as shown in \autoref{fig:SL_config} (left).
The projectors are assumed to be fixed with the camera at a certain relative rotation and position, \ie the projectors move with the camera.
Baseline lengths of the projectors are $20cm$ and $60cm$ from the camera, and field-of-view is $60^{\circ}$.
We synthesized each scene with different illumination conditions, such as normal illumination and no illumination.
The no illumination scenes are completely dark without ambient light and have no textural information except where the patterns are projected.
\autoref{fig:dataset_image} shows example images of the synthetic data with pattern projection.
%\vspace{-0.5cm}

\subsubsection{Evaluation on reconstruction and pose estimation accuracy}

\begin{figure}[t!]
    \centering
%    \begin{minipage}[b]{0.49\linewidth}
%        \centering
%        \includegraphics[width=\columnwidth]{figures/synthetic_qualitative_comparison_normal.pdf}
%        \small Normal illum.
%    \end{minipage}
%    \begin{minipage}[b]{0.49\linewidth}
%        \centering
%        \includegraphics[width=\columnwidth]{figures/synthetic_qualitative_comparison_black.pdf}
%        \small No illum.
%    \end{minipage}
    \includegraphics[width=\columnwidth]{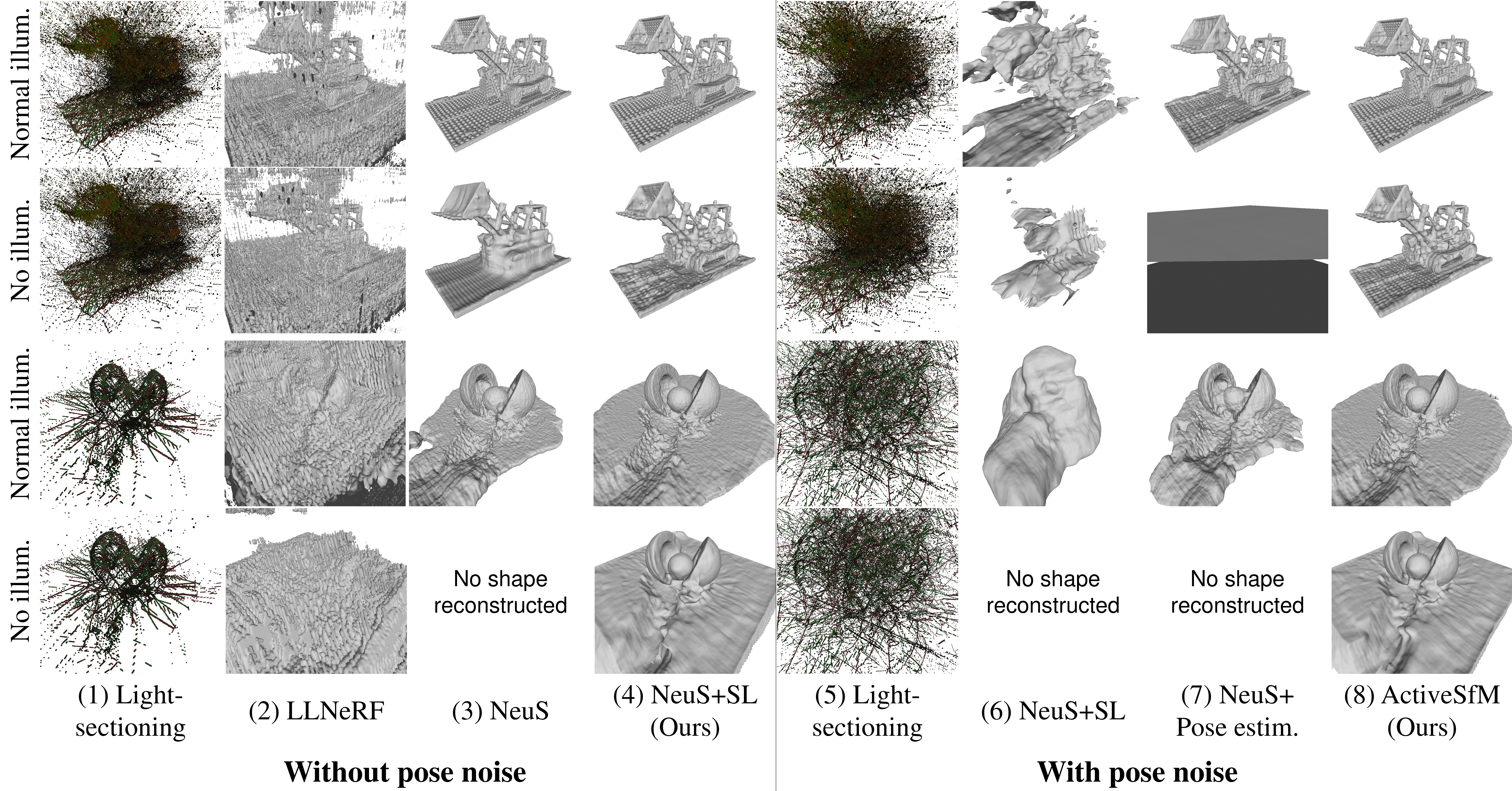}
    \caption{Reconstructed shapes with various methods. Qualitatively, ours achieved the best accuracy under no illumination with pose noise condition. Quantitative results are shown in \autoref{tab:synthetic_quantitative_comparison}. }
    \label{fig:synthetic_qualitative_comparison}
%    \vspace{-0.25cm}
\end{figure}

\begin{figure}[t!]
    \centering
    \begin{minipage}[b]{0.9\linewidth}
        \centering
        \includegraphics[width=\columnwidth]{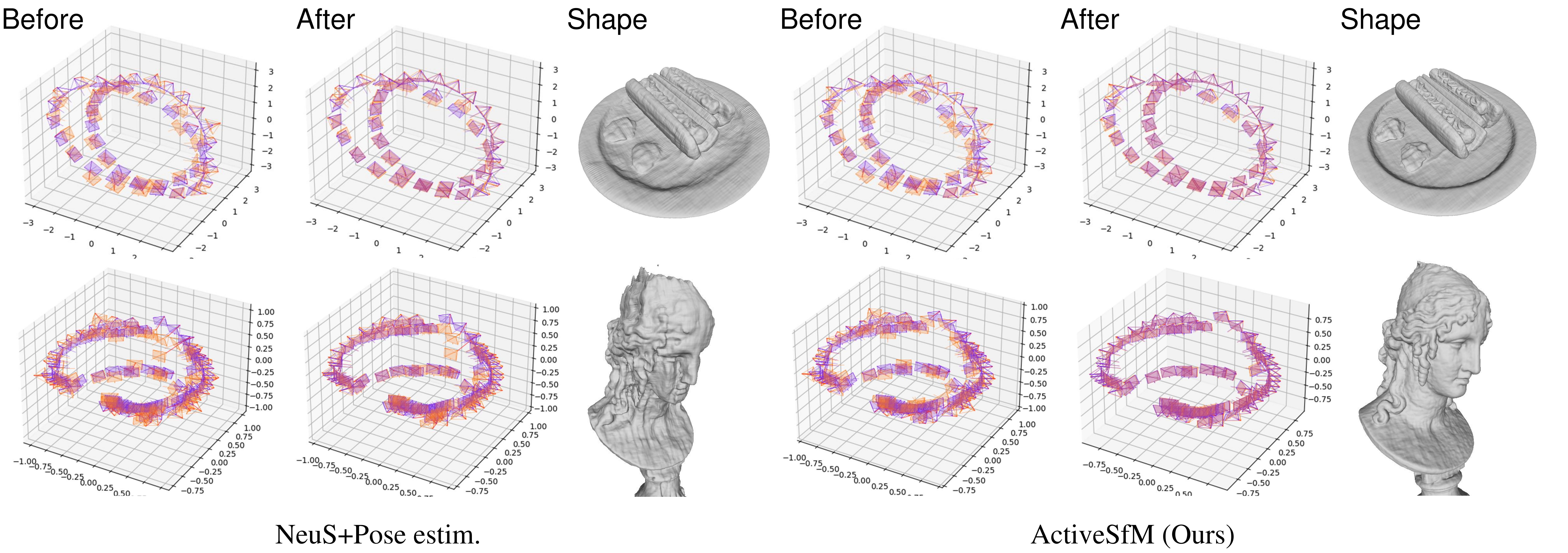}
    \end{minipage}
    \begin{minipage}[b]{0.9\linewidth}
        \centering
        \includegraphics[width=\columnwidth]{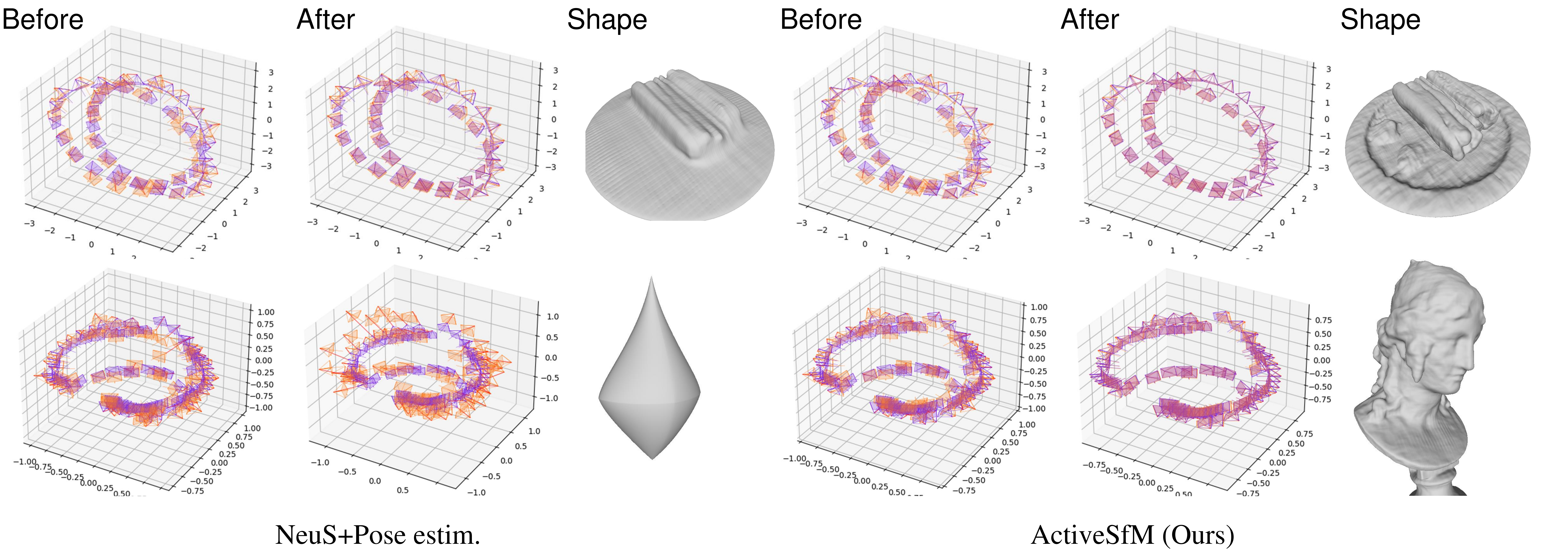}
    \end{minipage}
    \caption{Reconstruction and pose estimation results with the synthetic data. The left 3D map shows the initial system poses, and the right 3D map shows estimated system poses, for each scene. Blue frustums are the GT and orange frustum are the initial and estimated. \textbf{Top}: Normal illumination. \textbf{Bottom}: No illumination.}
    \label{fig:pose_estimation_comparison}
\end{figure}

\begin{table}[!ht]
    \centering
    \caption{Chamfer distances [mm] of the reconstructed shapes. Ours consistently outperforms comparative methods, especially when pose noise is added. "N/A" indicates no shape reconstructed. Legend: \textbf{Best}, \underline{Second best}.}
    \label{tab:synthetic_quantitative_comparison}
%    \vspace{-0.1cm}
    \scalebox{0.92}{
    \begin{tabular}{|ccc|cccc|cccc|}
    \hline
        Pose & \multirow{2}{*}{Illum} & \multirow{2}{*}{Method} & \multicolumn{4}{c|}{NeRF-Synthetic} & \multicolumn{4}{c|}{BlendedMVS} \\
        noise & & & Lego & Chair & Hotdog & Mic & Stone & Dog & Bear & Sculpture \\ \hline \hline
        \multirow{8}{*}{No} & \multirow{4}{*}{Normal} & Light-sectioning & 21.08 & \underline{16.15} & \textbf{14.00} & 26.24 & \textbf{17.40} & \textbf{6.02} & \textbf{6.95} & \textbf{4.84} \\
        ~ & ~ & LLNeRF & 19.73 & 26.40 & 33.44 & \underline{23.59} & 62.31 & 19.89 & 30.77 & 23.11 \\
        ~ & ~ & NeuS & \underline{11.79} & \textbf{9.25} & 28.08 & N/A & 44.40 & \underline{14.02} & \underline{14.15} & 23.20 \\
        ~ & ~ & NeuS+SL (Ours) & \textbf{11.62} & 20.63 & \underline{27.94} & \textbf{5.82} & \underline{30.92} & 17.83 & 30.84 & \underline{11.46} \\ \cline{2-11}
        ~ & \multirow{4}{*}{No} & Light-sectioning & \underline{21.08} & 16.15 & \textbf{14.00} & 26.24 & \textbf{17.40} & \textbf{6.02} & \textbf{6.95} & \textbf{4.84} \\
        ~ & ~ & LLNeRF & 23.14 & 32.36 & 40.30 & \underline{24.14} & 87.18 & 42.10 & 50.65 & 24.62 \\
        ~ & ~ & NeuS & 21.74 & \underline{12.12} & 32.27 & 58.73 & N/A & N/A & 86.05 & 12.52 \\
        ~ & ~ & NeuS+SL (Ours) & \textbf{14.97} & \textbf{8.51} & \underline{26.40} & \textbf{5.58} & \underline{35.35} & \underline{18.30} & \underline{25.63} & \underline{11.21} \\ \hline
        \multirow{8}{*}{Yes} & \multirow{4}{*}{Normal} & Light-sectioning & 39.88 & \underline{40.56} & 43.72 & \underline{39.90} & 96.97 & \underline{25.31} & 81.32 & \underline{20.80} \\
        ~ & ~ & NeuS+Pose estim. & \underline{13.73} & 50.97 & \underline{29.10} & N/A & \underline{59.51} & 28.04 & \textbf{23.17} & 28.57 \\
        ~ & ~ & NeuS+SL & 41.47 & 49.33 & 51.36 & 50.34 & 90.85 & 27.62 & 55.49 & 27.64 \\
        ~ & ~ & ActiveSfM (Ours) & \textbf{13.07} & \textbf{36.93} & \textbf{22.41} & \textbf{12.13} & \textbf{31.84} & \textbf{19.89} & \underline{23.40} & \textbf{10.75} \\ \cline{2-11}
        ~ & \multirow{4}{*}{No} & Light-sectioning & 39.88 & \underline{40.56} & \underline{43.72} & 39.90 & \underline{96.97} & \underline{25.31} & \underline{81.32} & \underline{20.80} \\
        ~ & ~ & NeuS+Pose estim. & 59.75 & 48.45 & 61.16 & \underline{22.69} & N/A & 29.31 & 81.40 & 32.26 \\
        ~ & ~ & NeuS+SL & \underline{35.04} & 44.76 & 52.76 & N/A & N/A & 56.93 & N/A & 70.74 \\
        ~ & ~ & ActiveSfM (Ours) & \textbf{18.50} & \textbf{18.60} & \textbf{28.47} & \textbf{6.37} & \textbf{35.36} & \textbf{21.67} & \textbf{22.99} & \textbf{11.20} \\ \hline
    \end{tabular}}
%    \vspace{-0.3cm}
\end{table}

\begin{table}[!ht]
    \centering
    \caption{Mean L1 errors of the estimated poses (rotation and translation). "N/A" indicates no shape reconstructed. Legend: \textbf{Best}.}
    \label{tab:pose_quantitative_comparison}
%    \vspace{-0.1cm}
    \scalebox{0.92}{
    \begin{tabular}{|ccc|cccc|cccc|}
    \hline
        \multirow{2}{*}{Illum} & \multirow{2}{*}{Metric} & \multirow{2}{*}{Method} & \multicolumn{4}{c|}{NeRF-Synthetic} & \multicolumn{4}{c|}{BlendedMVS} \\
        ~ & ~ & ~ & Lego & Chair & Hotdog & Mic & Stone & Dog & Bear & Sculpture \\ \hline \hline
        \multirow{4}{*}{Normal} & \multirow{2}{*}{Rot[$^\circ$]} & NeuS+Pose estim. & 0.84  & 3.90  & 1.50  & N/A & 0.60  & 3.02  & 0.96  & 2.40  \\
        ~ & ~ & ActiveSfM (Ours) & \textbf{0.33}  & \textbf{1.18}  & \textbf{1.29} & \textbf{0.88}  & \textbf{0.52}  & \textbf{0.76}  & \textbf{0.38}  & \textbf{0.67} \\ \cline{2-11}
        ~ & \multirow{2}{*}{Trans[$\%$]} & NeuS+Pose estim. & 4.25  & 25.63  & 9.61  & N/A & 1.37  & 7.83  & 1.53  & 4.44  \\
        ~ & ~ & ActiveSfM (Ours) & \textbf{1.26}  & \textbf{7.29}  & \textbf{5.59}  & \textbf{3.38}  & \textbf{0.09}  & \textbf{0.62}  & \textbf{0.43}  & \textbf{0.56}  \\ \hline
        \multirow{4}{*}{No} & \multirow{2}{*}{Rot[$^\circ$]} & NeuS+Pose estim. & 6.79  & 7.18  & 7.86  & 77.63  & N/A & 7.99  & 7.57  & 5.80  \\
        ~ & ~ & ActiveSfM (Ours) & \textbf{0.50}  & \textbf{1.07}  & \textbf{1.02}  & \textbf{0.63}  & \textbf{0.53}  & \textbf{1.01}  & \textbf{0.42}  & \textbf{0.90}  \\ \cline{2-11}
        ~ & \multirow{2}{*}{Trans[$\%$]} & NeuS+Pose estim. & 25.80  & 23.86  & 25.20  & 52.39  & N/A & 10.71  & 8.60  & 17.77  \\
        ~ & ~ & ActiveSfM (Ours) & \textbf{1.23}  & \textbf{3.70}  & \textbf{2.97}  & \textbf{2.32}  & \textbf{0.41}  & \textbf{1.78}  & \textbf{0.48}  & \textbf{0.71} \\ \hline
    \end{tabular}}
%    \vspace{-0.3cm}
\end{table}

First, we evaluated reconstruction and pose estimation accuracy of the proposed method.
We randomly added rotational and translational perturbations to the ground truth (GT) system poses, and measured mean L1 errors (rotation and translation) of the estimated poses to the GT poses and Chamfer distances of the reconstructed shapes to the GT shapes after training with the proposed method.
\knote{
To remove ambiguity on global rotation, translation and scaling, we applied ICP with scaling on the reconstructed shapes to the GT shapes.
}
As for perturbations, we used uniform noises with a constant value for rotation, and relative value to the mean baseline lengths between frames for translation, which are multiplied by baseline scaler as shown in \autoref{tab:datasets} since BlendedMVS images are not ordered.
The ranges of the uniform noises are $10^\circ$ for rotation, and $50\%$ for translation.
For more evaluations with different noise ranges, and the detailed procedure of noise addition, please refer to the supplementary material.

\autoref{tab:synthetic_quantitative_comparison} shows the results of the quantitative comparison, and \autoref{fig:synthetic_qualitative_comparison} shows the results of the qualitative comparison.
As shown in the results, the proposed method achieved the best accuracy under no illumination with pose noise condition.
Light-sectioning method is advantageous in low-light environment without pose noise, however, dense reconstruction is not possible and shapes are significantly corrupted with pose noise.
LLNeRF is capable of handling low-light illumination as well, however, it produced severe floating objects since it is not dedicated for shape reconstruction.
NeuS and NeuS+Pose estimation produced plausible shapes in some scenes, but they failed under no illumination scenario.
NeuS+SL could reconstruct accurate shapes only when there are no pose noise (Note that NeuS+SL is identical to ActiveSfM without pose noise).
Finally, ActiveSfM (Ours) successfully reconstructed accurate shapes in all scenes and scenarios, showing the feasibility of the proposed method.
As for the pose estimation accuracy, the proposed method achieved higher accuracy compared to NeuS+Pose estimation in all scenes, thanks to implicit depth supervision by pattern projection (\autoref{tab:pose_quantitative_comparison} and \autoref{fig:pose_estimation_comparison}).
%\vspace{-1.0cm}

\subsubsection{Evaluation on positional encoding}

\begin{figure}[t!]
    \centering
    \includegraphics[width=0.7\linewidth]{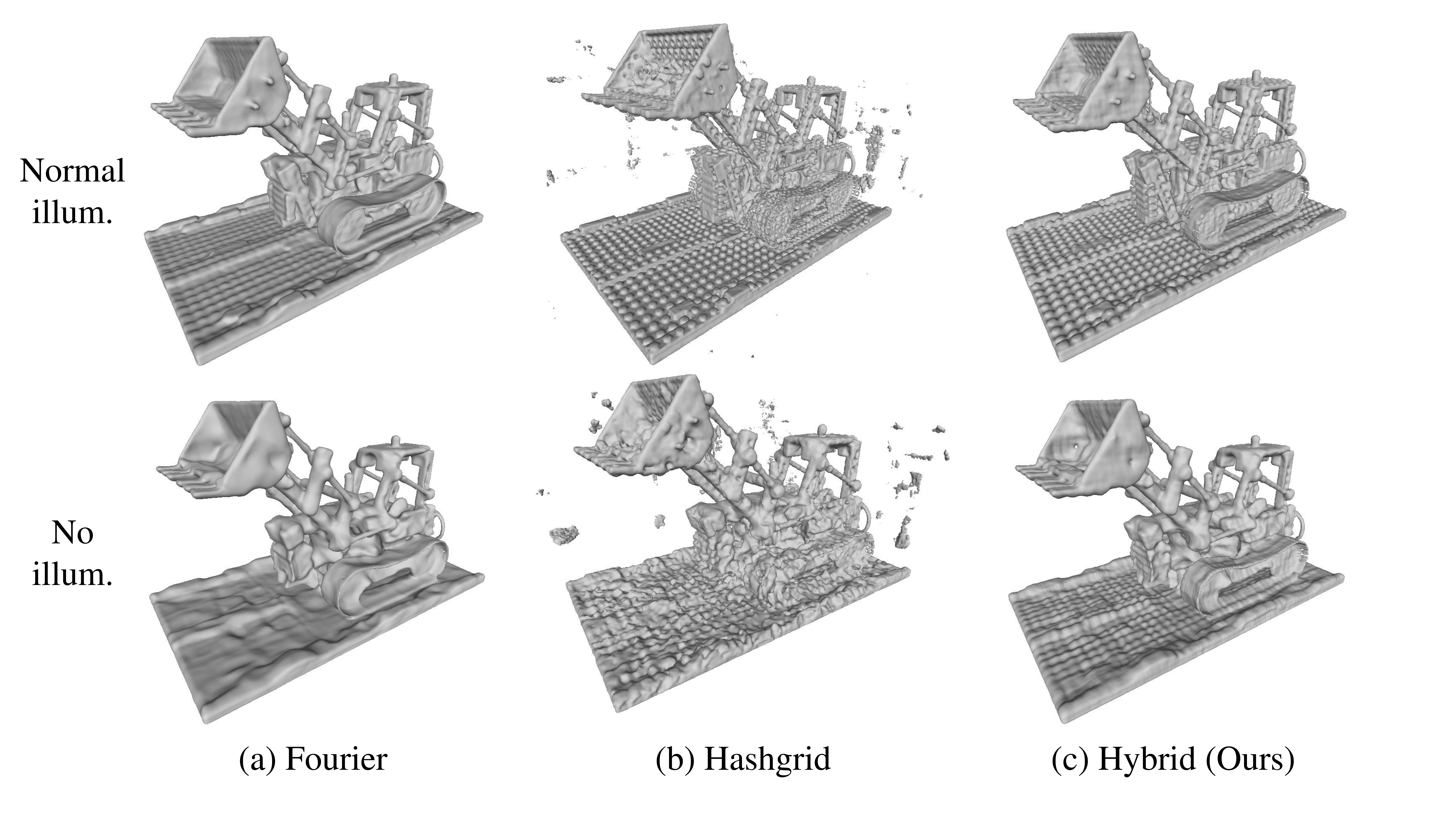}
    \caption{Reconstructed shapes with various positional encoding with pose estimation.}
    \label{fig:encoding_comparison}
%    \vspace{-0.5cm}
\end{figure}

\begin{table}[]
    \centering
    \small
    \caption{Results of the ablation study on positional encoding. \textbf{Best}, \underline{Second best}.}
    \label{tab:encoding_comparison}
    \begin{tabular}{|cc|ccc|}
        \hline
        Illum & Encoding & Rot[$^\circ$] & Trans[$\%$] & Shape[mm] \\ \hline \hline

        \multirow{3}{*}{Normal} & Fourier & \underline{0.53} & \textbf{1.04} & \underline{15.84} \\
        & Hashgrid & 4.73 & 34.98 & 45.63 \\
        & Hybrid & \textbf{0.33} & \underline{1.26} & \textbf{13.07} \\
%        & Full & 0 & 0 & 0 \\ 
        \hline

        \multirow{3}{*}{No} & Fourier & \underline{0.65} & \underline{1.37} & \underline{22.87} \\
        & Hashgrid & 1.31 & 1.45 & 23.58 \\
        & Hybrid & \textbf{0.50} & \textbf{1.23} & \textbf{18.50} \\
%        & Full & 0 & 0 & 0 \\

        \hline
    \end{tabular}
%    \vspace{-0.5cm}
\end{table}

Next, we conducted an ablation study on positional encoding.
We compared the reconstructed shapes with Fourier encoding, multi-resolution hash encoding (Hashgrid), and the proposed hybrid encoding (Hybrid).
%We used Lego scene of NeRF-Synthetic dataset with $10^\circ$ rotation noise and $50\%$ translation noise for the evaluation.

\autoref{fig:encoding_comparison} and \autoref{tab:encoding_comparison} show quantitative and qualitative comparisons of the proposed method against other positional encoding.
We observed that hybrid encoding consistently improves pose estimation and shape reconstruction in both normal and no illuminations.
In particular, observing the qualitative results, we can confirm major improvements in the fidelity of reconstructed shapes and reduction of floating objects which have little impact on the Chamfer distances.
%\vspace{-0.75cm}

\subsection{Evaluations with real data}
\label{ssec:real_evaluation}

\begin{figure}[t!]
    \centering
    \hspace{-0.3cm}
    \begin{minipage}[b]{0.13\linewidth}
        \centering
        \includegraphics[width=\columnwidth]{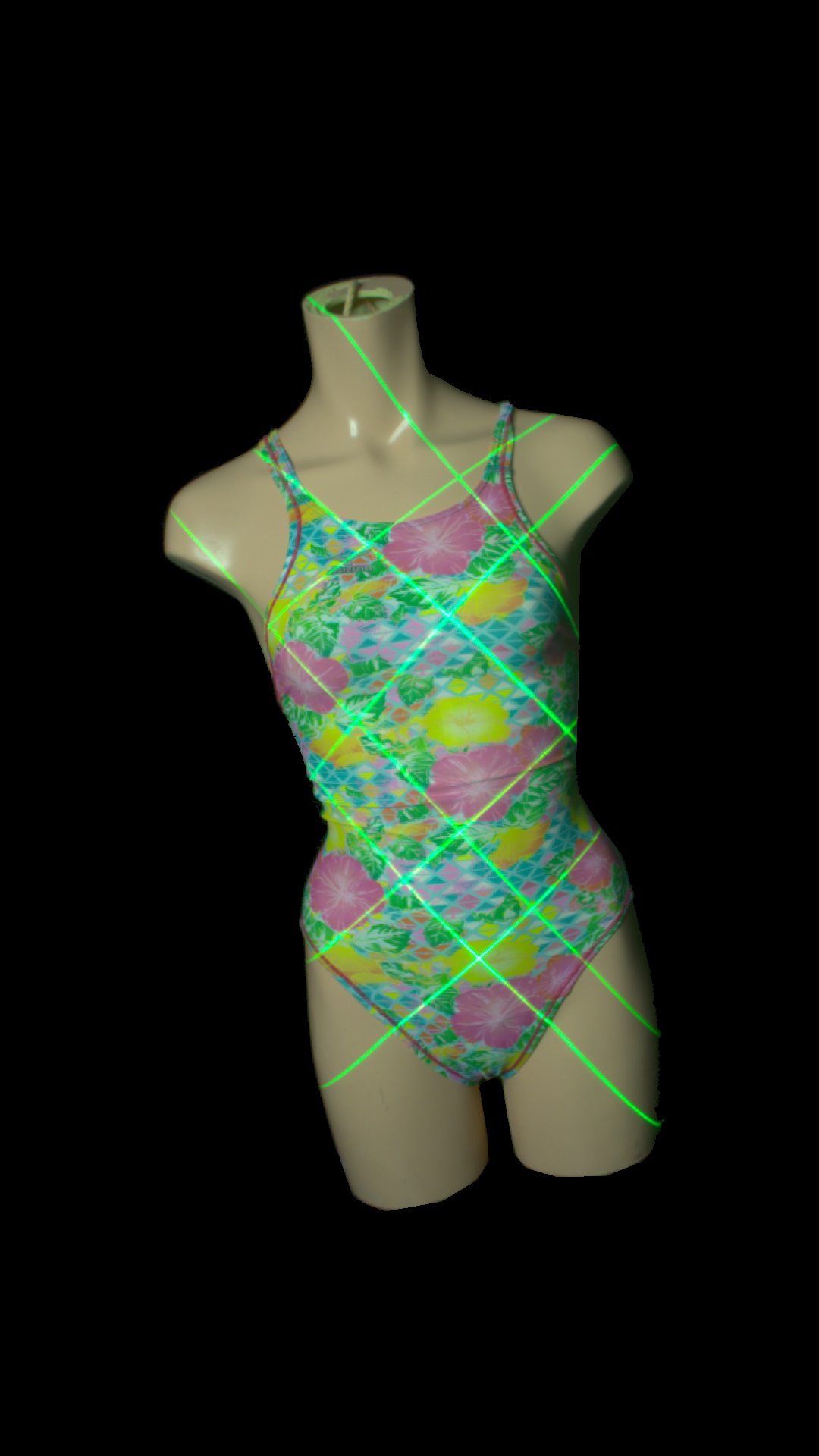}
    \end{minipage}
    \begin{minipage}[b]{0.13\linewidth}
        \centering
        \includegraphics[width=\columnwidth]{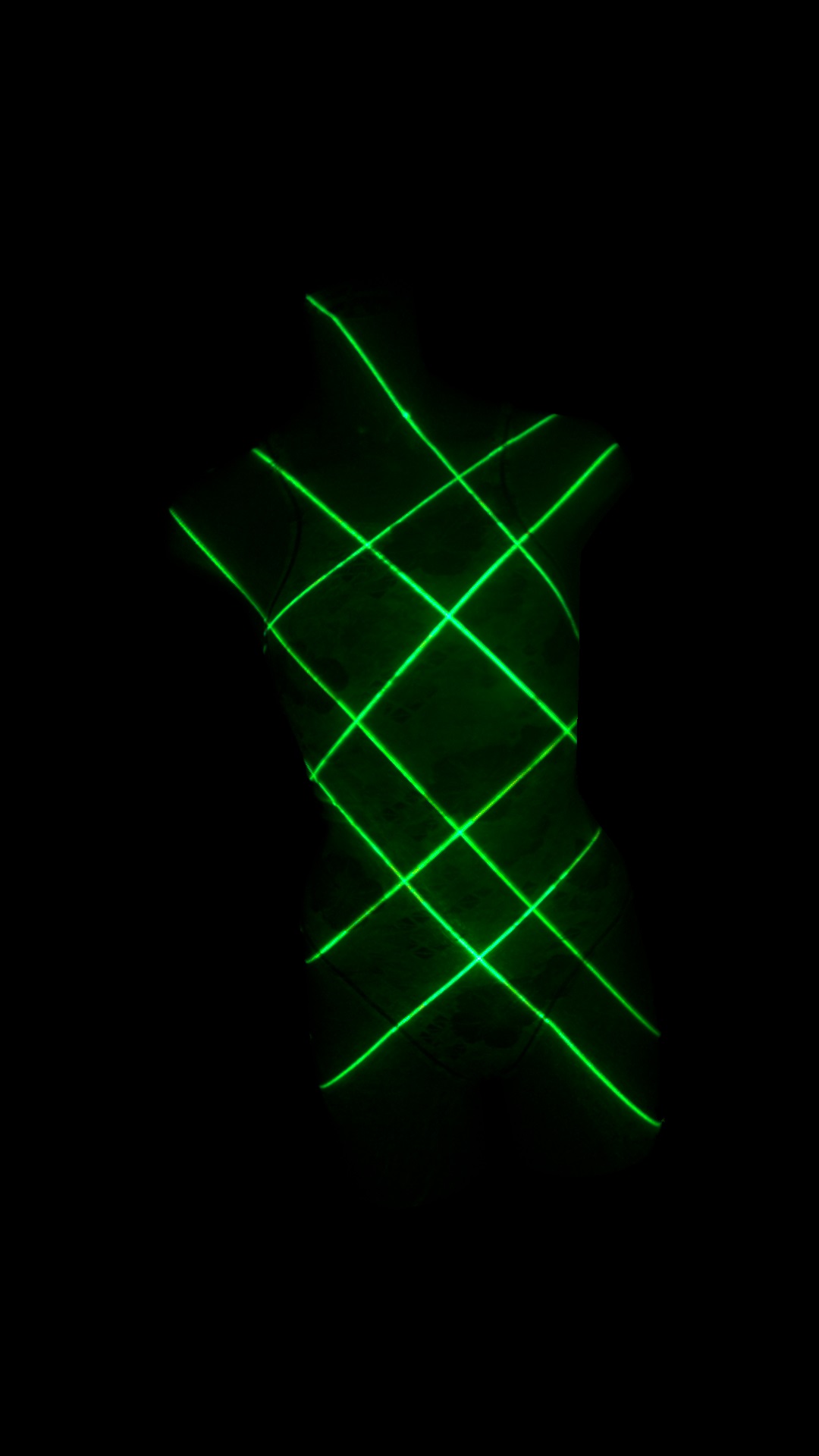}
    \end{minipage}
    \begin{minipage}[b]{0.36\linewidth}
        \centering
        \includegraphics[width=\columnwidth]{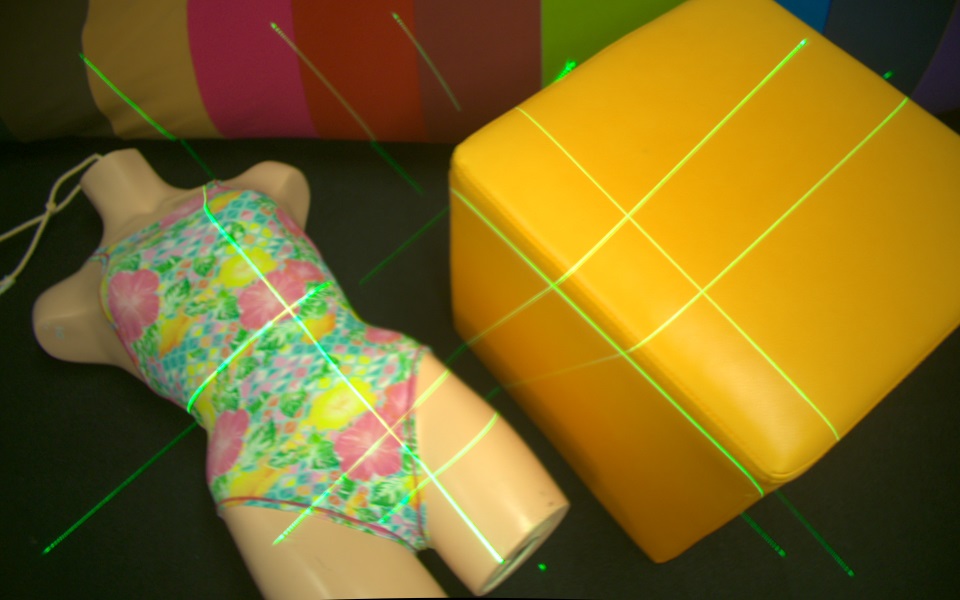}
    \end{minipage}
    \begin{minipage}[b]{0.36\linewidth}
        \centering
        \includegraphics[width=\columnwidth]{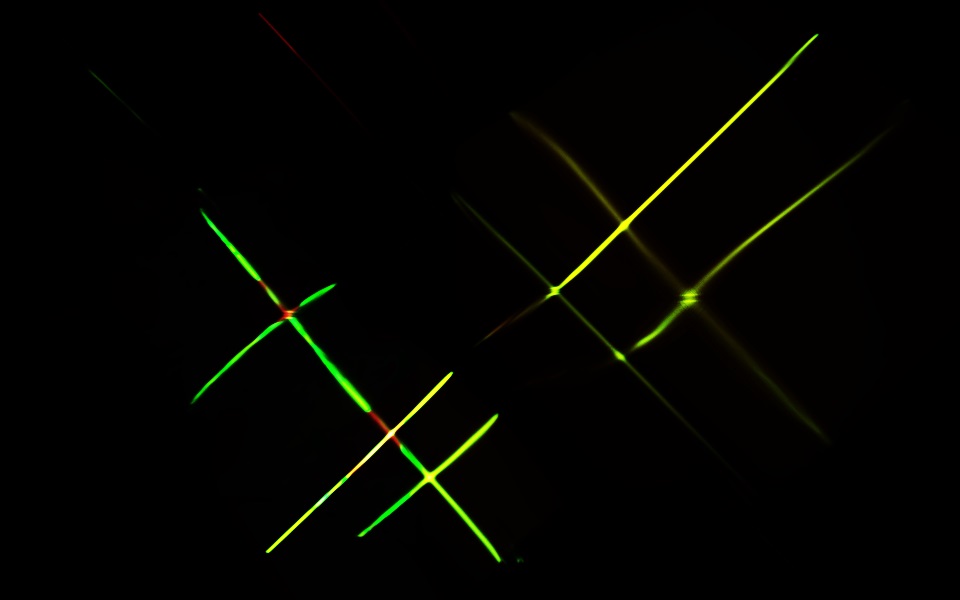}
    \end{minipage}
    \caption{Example of the real data (normal and no illumination). \textbf{Left}: Controlled sequence. \textbf{Right}: Freehand sequence. Contrast is enhanced for visualization.}
    \label{fig:real_image}
%    \vspace{-0.5cm}
\end{figure}

To confirm the feasibility of the proposed method in real scenes, we captured two real sequences with pattern projection by 4 cross laser projectors as shown in \autoref{fig:SL_config}, in both normal and no illuminations.
First, we used a turn table to rotate a mannequin (\ie, virtually moving the system around the mannequin), and captured images per $10^\circ$ in two illumination conditions (36 images per illumination condition, \autoref{fig:real_image}(left)).
Next, we used COLMAP to obtain the GT system poses relative to the mannequin using normal illumination (Note that we did not use the shapes obtained by COLMAP as GT).
To remove a scale ambiguity, we applied a light sectioning method~\cite{Nagamatsu2022ICPR} and adjusted the scale to fit the reconstruction from light sectioning into a point cloud from COLMAP.
Note that, since we captured images in two illumination conditions per viewpoint, we also have GT for the no illumination scene.
Then, we evaluated the proposed method using the sequences with rotational and translational pose noises.
As for the GT shape, we captured the same object using a ToF sensor and KinectFusion~\cite{KinectFusion}.

Similarly, we captured a scene with a freehand trajectory (78 images) to further evaluate the robustness of the method (\autoref{fig:real_image}(right)).
In the capturing process, the system was moved freely around the static scene by hand.
We captured a sequence in the normal illumination as a reference to obtain the GT poses by COLMAP and synthesized a no illumination sequence by reducing textures other than laser curves.
Then, we evaluated our method using the data by adding noise on rotation and translation of the poses as same as the controlled sequence.

\begin{figure}[t!]
    \centering
    \includegraphics[width=1.0\linewidth]{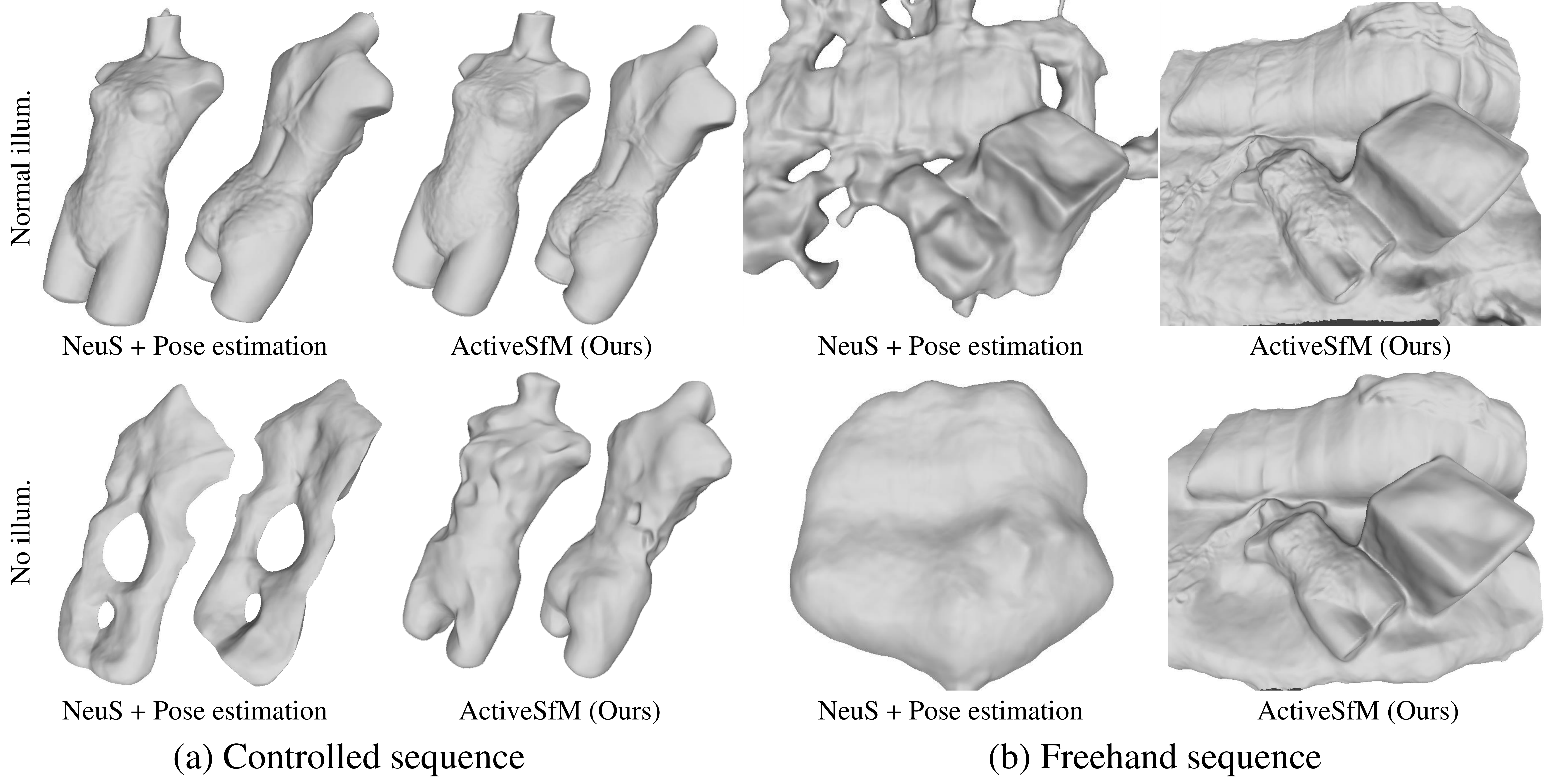}
    \caption{Reconstructed shapes on the real scenes with pose noise. NeuS + Pose estimation suffered from severe collapse except controlled sequence in normal illumination, while ours produced plausible shapes for all scenarios.}
    \label{fig:mannequin}
%    \vspace{-0.5cm}
\end{figure}

\begin{table}[!ht]
    \centering
    \caption{Qualitative comparison results of pose estimation and shape reconstruction on real data. Legend: \textbf{Best}, \underline{Second best}. ``-'' indicates rotational and translational error does not exist since there is no pose noise added.}
    \label{tab:mannequin}
    \scalebox{0.92}{
    \begin{tabular}{|ccc|ccc|cc|}
    \hline
        \multirow{2}{*}{Pose noise} & \multirow{2}{*}{Illum} & \multirow{2}{*}{Method} & \multicolumn{3}{c|}{Controlled sequence} & \multicolumn{2}{c|}{Freehand sequence} \\
        ~ & ~ & ~ & Rot[$^\circ$] & Trans[$\%$] & Shape[mm] & Rot[$^\circ$] & Trans[$\%$] \\ \hline \hline
        \multirow{8}{*}{No} & \multirow{4}{*}{Normal} & Light-sectioning & - & - & 13.33  & - & - \\
        ~ & ~ & LLNeRF & - & - & 15.93  & - & - \\
        ~ & ~ & NeuS & - & - & \textbf{9.39}  & - & - \\
        ~ & ~ & NeuS+SL(Ours) & - & - & \underline{9.55}  & - & - \\ \cline{2-8}
        ~ & \multirow{4}{*}{No} & Light-sectioning & - & - & \underline{13.33}  & - & - \\
        ~ & ~ & LLNeRF & - & - & 15.87  & - & - \\
        ~ & ~ & NeuS & - & - & 20.67  & - & - \\
        ~ & ~ & NeuS+SL(Ours) & - & - & \textbf{9.39}  & - & - \\ \hline
        \multirow{4}{*}{Yes} & \multirow{2}{*}{Normal} & NeuS+Pose estim. & 0.87  & 15.16  & \textbf{9.27}  & 2.22 & 52.77 \\
        ~ & ~ & ActiveSfM (Ours) & \textbf{0.70}  & \textbf{5.27}  & 9.37  & \textbf{1.87}  & \textbf{43.69}  \\ \cline{2-8}
        ~ & \multirow{2}{*}{No} & NeuS+Pose estim. & 4.94  & 742.14  & 24.41  & 2.18 & 64.60 \\
        ~ & ~ & ActiveSfM (Ours) & \textbf{2.02}  & \textbf{17.19}  & \textbf{12.29}  & \textbf{1.27}  & \textbf{18.89} \\ \hline
    \end{tabular}}
%    \vspace{-0.5cm}
\end{table}

\autoref{fig:mannequin} and \autoref{tab:mannequin} show the result.
The proposed method performed accurate shape reconstruction and pose estimation compared to NeuS+Pose estimation, especially under no illumination scenario.
Although the reconstructed shapes is a bit bumpy by our method under no illumination scenario,  the shapes and poses were mostly correctly estimated;
we consider it is mainly because of indirect illumination effects and it is important topic for future work.
From the table, translation error is considerably large under normal illumination compared to no illumination on the freehand sequence; we consider it is because small pose errors by COLMAP.
%pose had some errors.
%\vspace{-0.25cm}

\section{Conclusion}
\label{sec:conclusion}
%\vspace{-0.25cm}

In this paper, we proposed a simultaneous shape reconstruction and pose estimation method for SL systems, which we call Active SfM, using Neural SDF.
To achieve it, %enhance reconstruction quality under such a challenging condition, we also
we proposed a volumetric rendering pipeline for SL and introduced hybrid encoding for robust pose estimation and high-fidelity shape reconstruction. %and 2-stage training.
Experimental results show the proposed method can efficiently recover the scene geometry only from the information of projected patterns with rough initial poses in both synthetic and real dataset.
%s, and open the new door for autonomous exploration of in-the-wild environments.
%
As for the future work, we are interested in whether Neural SDF can cope with other challenging conditions, such as scattering, inter-reflection, occlusion, etc.
%diverse reflective properties, and so on.
\knote{
Concurrently, it is also important to address many other problems such as distortion due to refraction, attenuation, volumetric scattering to achieve an accurate deep sea vision system.
}

%\section*{ACKNOWLEDGMENT}
\subsubsection*{Acknowledgments}
%\vspace{-1mm}
%\vspace{3mm}
%\noindent \textbf{Acknowledgement}
This work was part supported by JST Startup JPMJSF23DR, ACT-X JPMJAX23C2 and JSPS/KAKENHI JP20H00611 and JP23H03439 in Japan.
%\vspace{-1mm}

\bibliographystyle{splncs04}
\bibliography{main}

\begin{thebibliography}{10}
\providecommand{\url}[1]{\texttt{#1}}
\providecommand{\urlprefix}{URL }
\providecommand{\doi}[1]{https://doi.org/#1}

\bibitem{alzuhiri2023imu}
Alzuhiri, M., Li, Z., Rao, A., Li, J., Fairchild, P., Tan, X., Deng, Y.: Imu-assisted robotic structured light sensing with featureless registration under uncertainties for pipeline inspection. NDT \& E International  \textbf{139},  102936 (October 2023)

\bibitem{chen_2020_autotuning}
Chen, W., Mirdehghan, P., Fidler, S., Kutulakos, K.N.: Auto-tuning structured light by optical stochastic gradient descent. In: The IEEE Conference on Computer Vision and Pattern Recognition (CVPR) (June 2020)

\bibitem{NeuralLambertian}
Cho, S.Y., Chow, T.: A neural-learning-based reflectance model for 3-d shape reconstruction. IEEE Transactions on Industrial Electronics  (2000)

\bibitem{ActivePhotometricStereo}
Clark, J.: Active photometric stereo. In: IEEE Computer Society Conference on Computer Vision and Pattern Recognition (1992)

\bibitem{Fernandez2012icip}
Fernandez, S., Salvi, J.: A novel structured light method for one-shot dense reconstruction. In: IEEE International Conference on Image Processing (2012)

\bibitem{Kawasaki2009:Article_Laser1276714555}
Furukawa, R., Kawasaki, H.: Laser range scanner based on self-calibration techniques using coplanarities and metric constraints. Computer Vision and Image Understanding  \textbf{113}(11),  1118--1129 (2009)

\bibitem{Furukawa_2022_WACV}
Furukawa, R., Mikamo, M., Sagawa, R., Kawasaki, H.: Single-shot dense active stereo with pixel-wise phase estimation based on grid-structure using cnn and correspondence estimation using gcn. In: Proceedings of the IEEE/CVF Winter Conference on Applications of Computer Vision (WACV). pp. 4001--4011 (January 2022)

\bibitem{Eikonal}
Gropp, A., Yariv, L., Haim, N., Atzmon, M., Lipman, Y.: Implicit geometric regularization for learning shapes. In: Proceedings of Machine Learning and Systems 2020, pp. 3569--3579 (2020)

\bibitem{Gu2020sensors}
Gu, F., Song, Z., Zhao, Z.: Single-shot structured light sensor for 3d dense and dynamic reconstruction. Sensors  \textbf{20}(4), ~1094 (2020)

\bibitem{huang2022hdr}
Huang, X., Zhang, Q., Feng, Y., Li, H., Wang, X., Wang, Q.: Hdr-nerf: High dynamic range neural radiance fields. In: Proceedings of the IEEE/CVF Conference on Computer Vision and Pattern Recognition. pp. 18398--18408 (2022)

\bibitem{Graycode}
Inokuchi, S., Sato, K., Matsuda, F.: Range imaging system for 3-d object recognition. In: International Conference on Pattern Recognition (1984)

\bibitem{KinectFusion}
Izadi, S., Kim, D., Hilliges, O., Molyneaux, D., Newcombe, R., Kohli, P., Shotton, J., Hodges, S., Freeman, D., Davison, A., Fitzgibbon, A.: Kinectfusion: Real-time 3d reconstruction and interaction using a moving depth camera. In: UIST '11 Proceedings of the 24th annual ACM symposium on User interface software and technology. pp. 559--568. ACM (October 2011), \url{https://www.microsoft.com/en-us/research/publication/kinectfusion-real-time-3d-reconstruction-and-interaction-using-a-moving-depth-camera/}

\bibitem{SCNeRF}
Jeong, Y., Ahn, S., Choy, C., Anandkumar, A., Cho, M., Park, J.: Self-calibrating neural radiance fields. In: ICCV (2021)

\bibitem{ProjectedTextureStereo}
Konolige, K.: Projected texture stereo. In: IEEE International Conference on Robotics and Automation (2010)

\bibitem{Chunyu2022DRSL}
Li, C., Hashimoto, T., Matsumoto, E., Kato, H.: Multi-view neural surface reconstruction with structured light. In: The British Machine Vision Conference (BMVC) (2022)

\bibitem{neuralangelo}
Li, Z., M\"uller, T., Evans, A., Taylor, R.H., Unberath, M., Liu, M.Y., Lin, C.H.: Neuralangelo: High-fidelity neural surface reconstruction. In: IEEE Conference on Computer Vision and Pattern Recognition ({CVPR}) (2023)

\bibitem{BARF}
Lin, C.H., Ma, W.C., Torralba, A., Lucey, S.: Barf: Bundle-adjusting neural radiance fields. In: IEEE International Conference on Computer Vision ({ICCV}) (2021)

\bibitem{mildenhall2022rawnerf}
Mildenhall, B., Hedman, P., Martin-Brualla, R., Srinivasan, P.P., Barron, J.T.: {NeRF} in the dark: High dynamic range view synthesis from noisy raw images. CVPR  (2022)

\bibitem{NeRF}
Mildenhall, B., Srinivasan, P.P., Tancik, M., Barron, J.T., Ramamoorthi, R., Ng, R.: Nerf: Representing scenes as neural radiance fields for view synthesis. In: ECCV (2020)

\bibitem{Parsa2018cvpr}
Mirdehghan, P., Chen, W., Kutulakos, K.N.: Optimal structured light a la carte. In: IEEE/CVF Conference on Computer Vision and Pattern Recognition (2018)

\bibitem{InstantNGP}
M\"uller, T., Evans, A., Schied, C., Keller, A.: Instant neural graphics primitives with a multiresolution hash encoding. ACM Trans. Graph.  \textbf{41}(4),  102:1--102:15 (Jul 2022). \doi{10.1145/3528223.3530127}, \url{https://doi.org/10.1145/3528223.3530127}

\bibitem{Nagamatsu2022ICPR}
Nagamatsu, G., Ikeda, T., Iwaguchi, T., Thomas, D., Takamatsu, J., Kawasaki, H.: Self-calibration of multiple-line-lasers based on coplanarity and epipolar constraints for wide area shape scan using moving camera. In: International Conference on Pattern Recognition (ICPR) (2022)

\bibitem{Nagamatsu2021IROS}
Nagamatsu, G., Takamatsu, J., Iwaguchi, T., Thomas, D., Kawasaki, H.: Self-calibrated dense 3d sensor using multiple cross line-lasers based on light sectioning method and visual odometry. In: IEEE/RSJ International Conference on Intelligent Robots and Systems (IROS) (2021)

\bibitem{CamP}
Park, K., Henzler, P., Mildenhall, B., Barron, J.T., Martin-Brualla, R.: Camp: Camera preconditioning for neural radiance fields. ACM Trans. Graph.  (2023)

\bibitem{NeRFMeshing}
Rakotosaona, M.J., Manhardt, F., Arroyo, D.M., Niemeyer, M., Kundu, A., Tombari, F.: Nerfmeshing: Distilling neural radiance fields into geometrically-accurate 3d meshes  (2023)

\bibitem{sarlin20superglue}
Sarlin, P.E., DeTone, D., Malisiewicz, T., Rabinovich, A.: {SuperGlue}: Learning feature matching with graph neural networks. In: CVPR (2020), \url{https://arxiv.org/abs/1911.11763}

\bibitem{schoenberger2016sfm}
Sch\"{o}nberger, J.L., Frahm, J.M.: Structure-from-motion revisited. In: Conference on Computer Vision and Pattern Recognition (CVPR) (2016)

\bibitem{NFSL}
Shandilya, A., Attal, B., Richardt, C., Tompkin, J., O’Toole, M.: Neural fields for structured lighting. In: IEEE/CVF Conference on Computer Vision and Pattern Recognition (2023)

\bibitem{PhaseShift}
Srinivasan, V., Liu, H.C., Halioua, M.: Automated phase-measuring profilometry of 3-d diffuse objects. Applied Optics  \textbf{23},  3105--3108 (1984)

\bibitem{DeepV2D}
Teed, Z., Deng, J.: Deep{V2D}: Video to depth with differentiable structure from motion. In: Proceedings of The International Conference on Learning Representations (ICLR) (2020)

\bibitem{tzathas2023}
Tzathas, P., Maragos, P., Roussos, A.: 3d neural sculpting (3dns): Editing neural signed distance functions. In: 2023 IEEE/CVF Winter Conference on Applications of Computer Vision (WACV). IEEE (January 2023)

\bibitem{wang2021nerf-sr}
Wang, C., Wu, X., Guo, Y.C., Zhang, S.H., Tai, Y.W., Hu, S.M.: Nerf-sr: High-quality neural radiance fields using supersampling. ACM International Conference on Multimedia  (2022)

\bibitem{llnerf}
Wang, H., Xu, X., Xu, K., Lau, R.W.: Lighting up nerf via unsupervised decomposition and enhancement. In: ICCV (2023)

\bibitem{NeuS}
Wang, P., Liu, L., Liu, Y., Theobalt, C., Komura, T., Wang, W.: Neus: Learning neural implicit surfaces by volume rendering for multi-view reconstruction. NeurIPS  (2021)

\bibitem{NeuS2}
Wang, Y., Han, Q., Habermann, M., Daniilidis, K., Theobalt, C., Liu, L.: Neus2: Fast learning of neural implicit surfaces for multi-view reconstruction. In: Proceedings of the IEEE/CVF International Conference on Computer Vision (ICCV) (2023)

\bibitem{NeRF--}
Wang, Z., Wu, S., Xie, W., Chen, M., Prisacariu, V.A.: Ne{RF}$--$: Neural radiance fields without known camera parameters. arXiv preprint arXiv:2102.07064  (2021)

\bibitem{yao2020blendedmvs}
Yao, Y., Luo, Z., Li, S., Zhang, J., Ren, Y., Zhou, L., Fang, T., Quan, L.: Blendedmvs: A large-scale dataset for generalized multi-view stereo networks. Computer Vision and Pattern Recognition (CVPR)  (2020)

\bibitem{VolSDF}
Yariv, L., Gu, J., Kasten, Y., Lipman, Y.: Volume rendering of neural implicit surfaces. In: Thirty-Fifth Conference on Neural Information Processing Systems (2021)

\bibitem{Zhang_2023_ICCV}
Zhang, J., Zhan, F., Yu, Y., Liu, K., Wu, R., Zhang, X., Shao, L., Lu, S.: Pose-free neural radiance fields via implicit pose regularization. In: Proceedings of the IEEE/CVF International Conference on Computer Vision (ICCV). pp. 3534--3543 (October 2023)

\bibitem{zhou2023high}
Zhou, J., Ji, Z., Li, Y., Liu, X., Yao, W., Qin, Y.: High-precision calibration of a monocular-vision-guided handheld line-structured-light measurement system. Sensors  \textbf{23}(14), ~6469 (2023)

\bibitem{SFMLearner}
Zhou, T., Brown, M., Snavely, N., Lowe, D.G.: Unsupervised learning of depth and ego-motion from video. In: Proceedings of the IEEE/CVF Conference on Computer Vision and Pattern Recognition (CVPR) (2017)

\end{thebibliography}

\end{document}